\documentclass[sigconf,nonacm]{acmart}
\AtBeginDocument{%
  }

\setcopyright{acmlicensed}
\copyrightyear{2025}
\acmYear{2025}
\acmDOI{XXXXXXX.XXXXXXX}
\acmConference[ACM KDD 2025]{2025 ACM SIGKDD International Conference on Knowledge Discovery and Data Mining}{August 03--07,
  2025}{Toronto, ON, Canada}
\acmISBN{978-1-4503-XXXX-X/2018/06}

\usepackage{algorithm,algpseudocode}
\usepackage[utf8]{inputenc} 
\usepackage[T1]{fontenc}    
\usepackage{hyperref}       
\usepackage{url}            
\usepackage{booktabs}       
\usepackage{amsfonts}       
\usepackage{nicefrac}       
\usepackage{microtype}      
\usepackage{graphicx}
\usepackage[export]{adjustbox}
\usepackage{todonotes}
\usepackage{tablefootnote}
\usepackage{bm}
\usepackage{subcaption}
\usepackage{diagbox}
\usepackage{booktabs}
\usepackage{multirow}
\usepackage{xspace}
\usepackage{cleveref}
\usepackage{paralist}
\usepackage{enumitem}

\newcommand{\ES}{\textbf{EvS}\xspace}

\begin{document}
\title{EBES: Easy Benchmarking for Event Sequences}

\author{Dmitry Osin}
\affiliation{%
  \institution{Skolkovo Institute of Science and Technology}
  \city{Moscow}
  \country{Russia}}
\email{d.osin@skoltech.ru}

\author{Igor Udovichenko}
\affiliation{%
  \institution{Skolkovo Institute of Science and Technology}
  \city{Moscow}
  \country{Russia}}
\affiliation{%
  \institution{Vega Institute Foundation}
  \city{Moscow}
  \country{Russia}}
\email{i.udovichenko@skoltech.ru}

\author{Egor Shvetsov}
\affiliation{%
  \institution{Skolkovo Institute of Science and Technology}
  \city{Moscow}
  \country{Russia}}
\email{e.shvetsov@skoltech.ru}

\author{Viktor Moskvoretskii}
\affiliation{%
  \institution{Skolkovo Institute of Science and Technology}
  \city{Moscow}
  \country{Russia}}
\affiliation{%
  \institution{HSE University}
  \city{Moscow}
  \country{Russia}}
\email{v.moskvoretskii@skoltech.ru}

\author{Evgeny Burnaev}
\affiliation{%
  \institution{Skolkovo Institute of Science and Technology}
  \city{Moscow}
  \country{Russia}}
\affiliation{%
  \institution{Artificial Intelligence Research Institute}
  \city{Moscow}
  \country{Russia}}
\email{e.burnaev@skoltech.ru}

\begin{abstract}
Event Sequences (\ES) refer to sequential data characterized by irregular sampling intervals and a mix of categorical and numerical features. Accurate classification of these sequences is crucial for various real-life applications, including healthcare, finance, and user interaction.
Despite the popularity of the \ES classification task, there is currently no standardized benchmark or rigorous evaluation protocol. This lack of standardization makes it difficult to compare results across studies, which can result in unreliable conclusions and hinder progress in the field.
To address this gap, we present \textbf{EBES}, a comprehensive benchmark for \ES classification with sequence-level targets. EBES features standardized evaluation scenarios and protocols, along with an open-source PyTorch library\footnote{Code is available at \url{https://github.com/On-Point-RND/EBES}} that implements 9 modern models. Additionally, it includes the largest collection of \ES datasets, featuring 10 curated datasets, including a novel synthetic dataset and real-world data with the largest publicly available banking dataset. The library offers user-friendly interfaces for integrating new methods and datasets.
Our benchmarking results highlight the unique properties of \ES compared to other sequential data types, provide a performance ranking of modern models—with GRU-based models achieving the best results—and reveal the challenges associated with robust \ES learning.
The goal of EBES is to facilitate reproducible research, expedite progress in the field, and increase the real-world impact of \ES classification techniques.
\end{abstract}

\maketitle


\section{Introduction}
\label{sec:intro}
The world in which we live is constantly changing~\cite{laertius1925lives}.
We continuously collect and analyze data to understand and navigate this dynamic environment.
This ongoing data collection helps capture the evolving nature of reality and can be captured in sequential datasets, which can be further analyzed or used for modeling.

Sequential data encompass a wide array of formats, from texts and videos to financial transaction logs and physical measurements. 
In this work, we focus on a specific subset of sequential data, termed \textbf{Event Sequences (\ES)}. 
These sequences consist of data points, each of which can be described by both categorical and numerical attributes. Every data point in an \ES is linked to a specific timestamp, and these points are arranged into sequences according to their timestamps.
This specific  structure is prevalent across various domains, including but not limited to healthcare~\citep{reyna2020early, goldberger2000physiobank}, ecology~\citep{clark2004population}, e-commerce~\citep{zhao2023percltv}, and finance~\citep{bazarova2024universal, mbd}.

In the real world, many important problems involve \ES classification, where the goal is to attribute a target label to entire sequences rather than individual points. This type of classification is crucial in various domains, such as mortality prediction~\citep{johnson2016mimic}, churn prediction~\citep{jain2021telecom}, predictive business process monitoring~\citep{marquez2017predictive}, and fraud detection~\citep{xie2022time}.

Despite its popularity \ES classification lacks established benchmark and rigorous evaluation protocol.
It leads to studies reporting inconsistent performance metrics~\citep{li2024time, chowdhury2023primenet, shukla2021multitime, horn2020set}, making it hard to compare classification models.
Manual hyperparameter tuning may lead to test leakage and biased estimation of performance~\citep{lones2021avoid}.

To address the mentioned problems we present EBES, a benchmark tailored for \ES classification.
In summary, our contributions are as follows:
\begin{enumerate}
    \item A rigorous evaluation protocol with statistically proven results and a curated list of datasets specifically designed for the classification of \ES.
    \item An open-source PyTorch library that implements the described evaluation protocol. This library is user-friendly and features unified interfaces, enabling the seamless addition of new methods and datasets in a plug-in fashion, without the need for individual adaptations.
    \item Evaluation results highlighting the need to treat \ES separately from closely related types of sequential data such as time series.
    \item A comprehensive empirical analysis that provides valuable insights into the properties of real-world \ES datasets and models developed for sequential data.
\end{enumerate}

The paper is organized as follows.
We formally define \ES classification in Section~\ref{sec:background}.
The evaluation methodology is described in detail in Section~\ref{sec:methodology}.
In Section~\ref{sec:benchmark}, we list the datasets and methods selected for the benchmark.
We present an empirical study of our benchmark in Section~\ref{sec:experiments}.


\section{Background}\label{sec:background}
\textbf{Definition.} Sequential data classification involves datasets comprising pairs $(S_i, y_i)$, where $S_i$ represents a sequence and $y_i$ is the classification target attributed to the entire sequence.
Each sequence $S_i = \{(x_{i}^j, t_{i}^j)\}_{j=1}^{n_i}$ consists of feature sets $x_{i}^j$ and corresponding timestamps $t_{i}^j$. 
The nature of these sequences can be categorized based on the regularity of timestamps and the type of features (see Figure~\ref{fig:types}):

\begin{figure}
    \centering
    \subfloat[Regularly Sampled Time Series (TS).]{%
        \label{fig:types_ts}%
        \includegraphics[width=0.2\textwidth,valign=t]{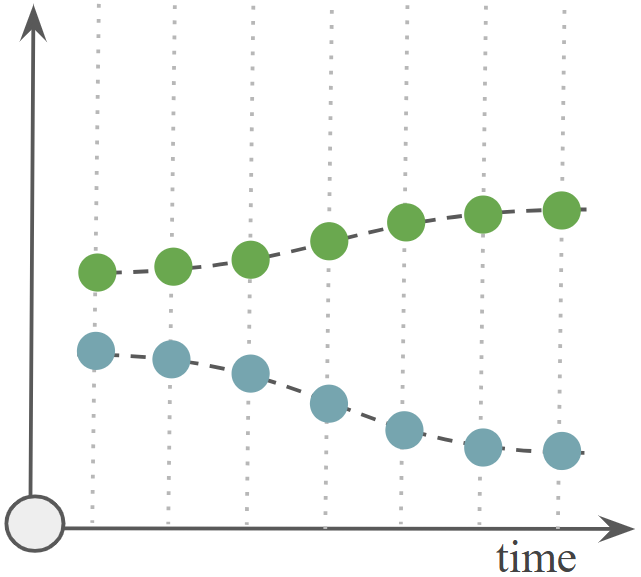}%
    }
    \hfill
    \subfloat[Continuous \ES with missing values]{%
        \label{fig:types_ists}%
        \includegraphics[width=0.2\textwidth,valign=t]{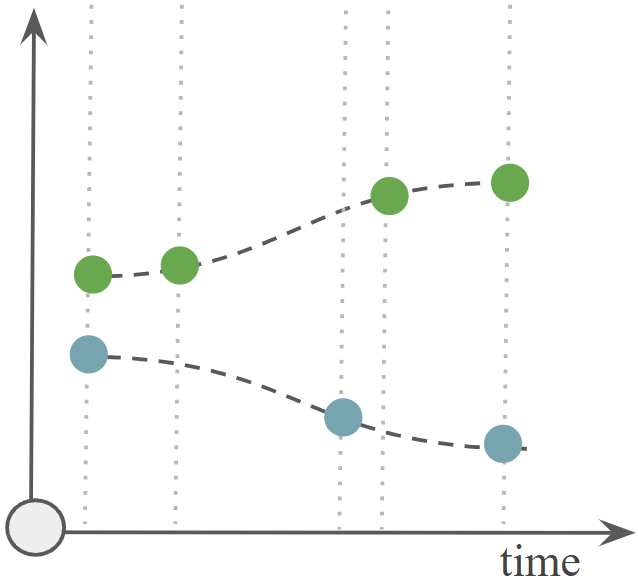}%
    }
    \hfill
    \subfloat[A stream of discrete events, usually, modeled by Temporal Point Process (TPP).]{%
        \label{fig:types_tpp}%
        \includegraphics[width=0.2\textwidth,valign=t]{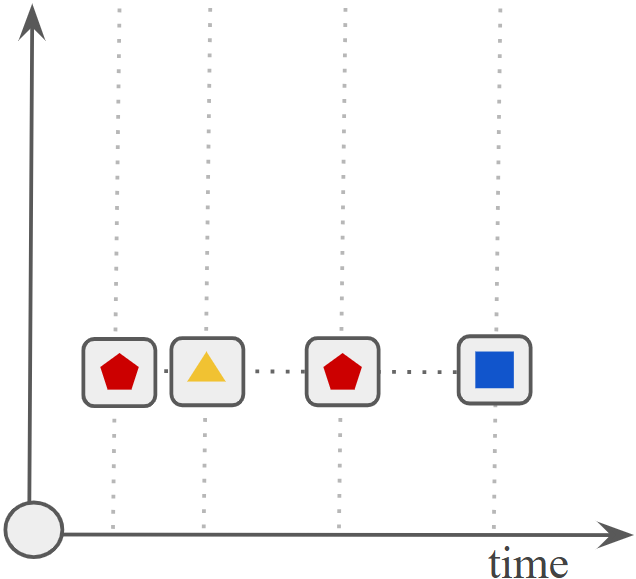}}%
    \hfill
    \subfloat[Discrete \ES with 2 numerical and 1 categorical features.]{%
        \label{fig:types_es}%
        \includegraphics[width=0.2\textwidth,valign=t]{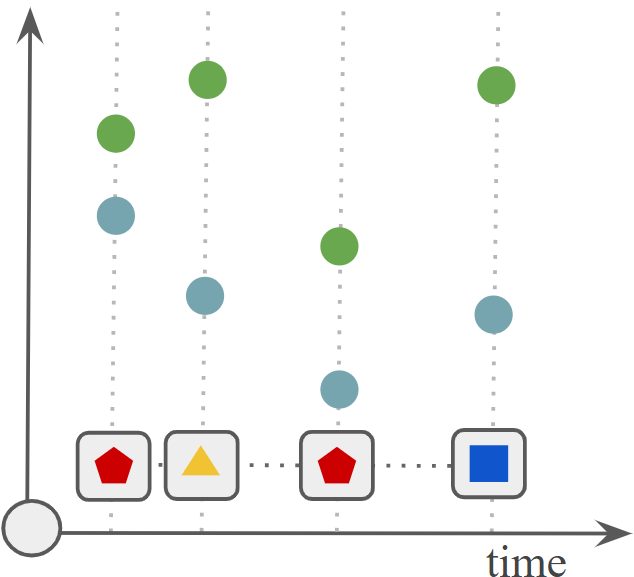}%
    }
    \caption{Categorization of sequential data. Green and blue dots indicate numerical features, while different shapes denote categorical features. For TS and Continuous \ES, an underlying process is present. For Discrete \ES, no underlying process exists, making interpolation between neighboring points meaningless.}
    \label{fig:types}
\end{figure}

\begin{itemize}[leftmargin=*, itemsep=0pt, parsep=0pt]
    \item \textbf{Time Series:} When timestamps $t_{i}^j$ are \textit{regular} and features $x_{i}^j$ are exclusively numerical, the sequence is classified as a time series.
    \item \textbf{Streams of discrete events:} When timestamps $t_{i}^j$ are \textit{irregular} and features $x_{i}^j$ are single categorical values, the sequence is modeled as a stream of discrete events, typically using Temporal Point Processes (TPP).
    \item \textbf{Event Sequences:} Our focus is on event sequences, characterized by \emph{irregular} timestamps and diverse feature types. These sequences can be further subdivided:
    \begin{itemize}[leftmargin=*, itemsep=0pt, parsep=0pt]
        \item \textbf{Continuous Event Sequences:} If all features $x_{i}^j$ are snapshots of an underlying continuous process, the data is termed a \textbf{continuous \ES}. These sequences often require interpolation techniques for analysis~\cite{shukla2018interpolation, shukla2021multitime, chowdhury2023primenet}.
        \item \textbf{Discrete Event Sequences:} If the interpolation between neighboring points in a sequence is meaningless (for example, card transactions), the data is classified as a \textbf{discrete \ES}.
    \end{itemize}
\end{itemize}


\section{Methodology}\label{sec:methodology}

In this section, we describe the evaluation methodology.
The goal is to enable a robust and rigorous comparison of methods, ensuring accurate conclusions.

\subsection{Data prepossessing}
In our work, we follow common practices whenever possible to prevent data preprocessing from affecting model evaluation. 
For ease of extensibility, we convert all datasets into a single format and release scripts that perform the conversion. 
Our data preprocessing includes:
\begin{itemize}[leftmargin=10pt]
    \item Applying a logarithm to fat-tailed variables, which are selected manually according to \citep{babaev2022coles};
    \item Rescaling time points to ensure the time range of all sequences falls within $[0,1]$;
    \item For missing values, we propagate them forward for datasets with continuous \ES based on results from~\cite{che2018recurrent}, and impute with constants for others.
\end{itemize}
EBES's data preprocessing is highly flexible and customizable, as it is defined by a single YAML config.

\begin{figure}[t]
    \includegraphics[width=0.45\textwidth]{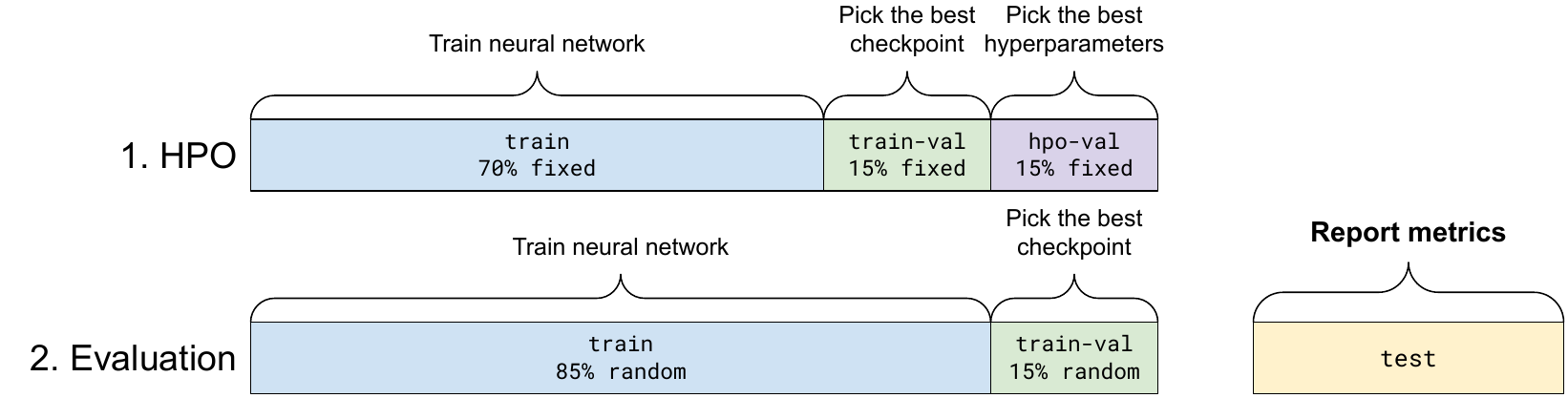}
    \caption{Data splits and their usage in our evaluation procedure. For each seed, the training sample is randomly divided into \texttt{train}, \texttt{train-val}, and \texttt{hpo-val} sets, while the \texttt{test} set is separated only once during data preprocessing.}
    \label{fig:methodology}
\end{figure}

In the preprocessing step, we split the data into full-train and test subsets.
For datasets with an established test set, we use that.
For datasets without a predefined test set, we employ time-based splitting when possible, reserving the last 20\% of sequences for testing.
When time-based splitting is not feasible, we randomly select 20\% of the sequences for testing.

\subsection{Hyperparameter optimization}\label{sec:mod_eval}
Hyperparameters are a fundamental aspect of machine learning that directly impacts model performance.
However, the procedure of hyperparameter tuning is rarely described.
Therefore, this becomes a source of non-reproducibility~\cite{arnold2023role, gundersen2022sources}. 
Moreover, manual hyperparameter tuning can lead to the leakage of the test set into the training procedure and performance~\cite{lones2021avoid};
at the same time, testing different hyperparameter values is necessary to find a model that generalizes well~\cite{gundersen2022sources}. 

Hyperparameter optimization (HPO) is at the core of our methodology.
For HPO we use Optuna~\citep{akiba2019optuna} Tree-structured Parzen Estimator (TPE), as its effectiveness has been proven on real-world tasks \cite{klyuchnikov2022bench}.
For each model and each dataset we allocate an HPO budget of 210 train runs capping the total compute at 18 NVidia A100 GPU-days.
The best hyperparameters are used to evaluate the model.
We emphasize that the procedure of HPO is fully formalized ensuring the test set is not leaked implicitly into the HPO procedure.
To train the model during HPO runs, we use only the full-train subset. This subset is split into three parts:
\begin{itemize}[leftmargin=*, itemsep=0pt, parsep=0pt]
\item \texttt{train}: Used for training the models.
\item \texttt{train-val}: Used for the early stopping procedure and checkpoint selection. Training is stopped if the target metric does not improve after several epochs and exceeds the patience limit.
\item \texttt{hpo-val}: Used for model evaluation during HPO.
\end{itemize}

These splits remain fixed during the HPO.
Both \texttt{train-val} and \texttt{hpo-val} take 15\% from the initial train dataset.
See Figure ~\ref{fig:methodology} for clarification.

\subsection{Model evaluation}

For model evaluation, we retrain the model with the best hyperparameters 20 times, each time using a different random seed. 
For each run, we \emph{randomly} split the full-train data subset into a \texttt{train} set (85\%) and a \texttt{train-val} set (15\%). 
The \texttt{train} set is used to train the neural network weights, while the \texttt{train-val} set is used for early stopping and selecting the best checkpoint. 
This approach is known in the literature as Monte Carlo cross-validation~\citep{xu2001monte}.

We report the mean and standard deviation of the target metric computed on the test set across these 20 runs.
To rank the models rigorously we perform Mann-Whitney tests~\cite{mann1947test} with Benjamini–Hochberg correction~\cite{benjamini1995controlling}.



\section{Benchmark}\label{sec:benchmark}

In this section, we describe the datasets and the methods we selected of our benchmark.





\subsection{Datasets}
We curated datasets from each domain: Time Series, Continuous \ES and Discrete \ES to illustrate distinctions between them.

\begin{itemize}[leftmargin=*, itemsep=0pt, parsep=0pt]
\item \textbf{Discrete \ES:} We selected three widely-used datasets: AGE, Retail, and Taobao, based on previous studies~\cite{shukla2021multitime, udovichenko2024seqnas, babaev2022coles, moskvoretskii2024selfsupervised, bazarova2024universal}. 
We also introduce two datasets to the broad community.
The first is the BPI17 dataset~\cite{bpi17data}, widely used in the subject field yet not well explored in general works on event sequences modeling.
The dataset comprises the logs of the business process of negotiating a loan request.
The second dataset is one of the largest event sequence datasets, MBD~\cite{mbd}.
It contains multi-modal data about the bank clients.
We take only the transaction log from the dataset.

\item \textbf{Continuous \ES:}
We utilized two medical datasets, PhysioNet 2012 and MIMIC-III. Additionally, we included a synthetic Pendulum dataset, which was specifically designed not only as an example of continuous \ES but also to validate the importance of temporal dynamics and assess how models capture sequential properties. 

\item \textbf{Time Series:}
To highlight the distinctions of \ES we also chose one multivariate dataset (ArabicDigits) and one univariate dataset (ElectricDevices).
\end{itemize}

Our dataset selection spans various domains, including two medical, three banking, one retail, one synthetic, and one business process mining dataset, reflecting diverse complexities and difficulties. We also included datasets of varying sizes to address scalability challenges. All datasets are open-access to promote reproducibility and collaboration and we welcome contributions from other domains to enrich our collection. Adding new datasets is straightforward: simply convert the dataset into the correct format and prepare a YAML configuration file following the provided examples. Table~\ref{table:datasets} presents statistics for each dataset, and Appendix~\ref{sec:app_data} provides detailed descriptions.

\begin{table*}[t]
\centering
\caption{Statistics of sequential datasets used in our benchmark. The statistics are calculated on the train set if not specified otherwise. }
\label{table:datasets}
\resizebox{\textwidth}{!}{\begin{tabular}{lrrrrrrrrrr}
\toprule
Dataset                     & Domain & \# classes & Class balance, \%      & Target         & Category             \\
\midrule
ElectriDevices               & Time Series                & 7          & 27 / 25 / 17 / 10 / 8 / 8 / 6 & Device type & Engineering         \\
ArabicDigits                 & Time Series                & 10         & balanced                & Digit & Motion              \\
PhysioNet 2012               & Continuous \ES    & 2          & 86 / 14                & Mortality      & Medical             \\
MIMIC-III                    & Continuous \ES    & 2          & 90 / 10                & Mortality      & Medical             \\
Pendulum                     & Continuous \ES    & 10         & balanced                & Air resistance & Physical (synth.)   \\
AGE                          & Discrete \ES    & 4          & 25 / 25 / 25 / 25      & Age group      & Transactions        \\
Retail                           & Discrete \ES    & 4          & 27 / 21 / 27 / 24      & Age group      & Transactions        \\
MBD                          & Discrete \ES  & 4 × 2      & 99.7 ± 0.2 / 0.3 ± 0.2 & Purchase items      & Transactions        \\
Taobao                       & Discrete \ES    & 2          & 43 / 57                & Purchase event       & E-commerce          \\
BPI17                        & Discrete \ES    & 2            & 70 / 30                & Offer acceptance & Business process mining  \\
\midrule
                             & \# seq. (train / test) & \# points (train / test) & \# points per seq. (mean ± std) & \# cat. features & \# num. features \\
\midrule
ElectriDevices               & 9k / 8k                     & 857k / 740k                     & 96 ± 0                           & 0                        & 1 \\
ArabicDigits                 & 7k / 2k                     & 263k / 87k                      & 40 ± 9                           & 0                        & 13 \\
PhysioNet 2012               & 4k / 4k                     & 299k / 299k                     & 75 ± 23                          & 3                        & 38                     \\
MIMIC-III                    & 45k / 11k                   & 2.7m / 657k                     & 58 ± 93                          & 1                        & 10                     \\
Pendulum                     & 80k / 20k                   & 2.5m / 631k                     & 32 ± 9                           & 0                        & 2                      \\
AGE                          & 24k / 6k                    & 21m / 5.3m                      & 881 ± 125                        & 1                        & 1                      \\
Retail                           & 319k / 80k                  & 37m / 9.1m                      & 114 ± 103                        & 7                        & 9                      \\
MBD                          & 7.4m / 1.8m                 & 156m / 39m                      & 21 ± 435                         & 11                       & 1                      \\
Taobao                       & 18k / 9k                    & 5.1m / 2.8m                     & 280 ± 387                        & 2                        & 0                      \\
BPI17                        & 34k / 9k                    & 444k / 119k                     & 13 ± 9                           & 7                        & 5                       \\
\bottomrule
\end{tabular}}
\end{table*}

\subsection{Models}\label{sec:models}

We have carefully curated diverse models and approaches previously applied to \ES classification tasks. In this section, we first list models from two broad categories: general-purpose sequential data models and models specialized for certain domains. Following this, we describe the common structure shared by all models.

\subsubsection{General Sequential Data Models}
This category includes widely-used architectures designed for handling sequential data.
Specifically, we include \textbf{GRU}~\citep{chung2014empirical}, \textbf{Mamba}~\citep{gu2023mamba}, and \textbf{Transformer}~\citep{vaswani2017attention}, all trained in a supervised manner.
These models are strong baselines due to their proven effectiveness across sequential data domains.
As a simple baseline, we evaluate a multi-layer perceptron (\textbf{MLP}) --- a 3-layer fully-connected neural network that takes aggregated events in a sequence as input.

\subsubsection{Models Specialized for Certain Domain.}
The second category comprises methods designed explicitly for certain domains of sequential data.
Among these, \textbf{CoLES}~\cite{babaev2022coles} and \textbf{MLEM}~\cite{moskvoretskii2024mlem} leverage contrastive learning or a combination of contrastive and generative strategies for unsupervised pre-training, followed by fine-tuning on downstream classification tasks.
In addition, we include models tailored for continuous \ES data. 
The \textbf{mTAND} architecture, introduced in \citep{shukla2021multitime}, is an attention-based model designed to interpolate and classify irregularly sampled time series with missing values.
Building on this, \textbf{PrimeNet} \citep{chowdhury2023primenet} extends mTAND by incorporating time-sensitive contrastive learning and data reconstruction tasks during pre-training.
To assess the performance of state-of-the-art methods in multivariate time series classification, we also include \textbf{\mbox{ConvTran}}~\cite{foumani2024improving}. 
Refer to Appendix~\ref{sec:app_models} for a detailed model description.

\subsubsection{Common Model Structure.}
Most selected methods share a common four-step structure, enabling systematic comparisons of design choices such as event time processing and sequence aggregation. This structure is reflected in the source code of EBES, making it easier to implement and test new models. Below, we outline these steps:
\begin{enumerate}[leftmargin=*]
    \item \textbf{Preprocessing block.} 
        This step transforms raw event sequences, described by categorical and real-valued features, into vector representations suitable for neural networks. The block encodes categorical features using embedding layers. For numerical features, it applies batch normalization and projects each feature individually into a larger space using a linear transformation.

        Two models explicitly process timestamps through specialized mechanisms. To ensure that all models can effectively utilize temporal information, regardless of their inherent design, timestamps can be concatenated with other event features in various ways: as absolute times, as time differences between consecutive events, or not concatenated at all. This approach treats time as just another feature describing an event.
        
    \item \textbf{Encoder block.}
        The encoder block maps the sequence of event vectors into another sequence of latent representations. 
        The architectures differ most significantly here, with each model employing its unique mechanism for capturing temporal dependencies.
    \item \textbf{Aggregation of latent representations.}
        The block aggregates a sequence of latent vectors into a single vector. 
        Common strategies include taking the last vector in the latent sequence or computing an average across the entire sequence.
    \item \textbf{Classification Head.}
        The final block predicts class logits from the aggregated representation, completing the classification process.
\end{enumerate}
This unified structure facilitates a fair comparison of different design choices across models and datasets, highlighting the impact of specific architectural decisions on \ES classification performance.

\section{Results and Analysis}\label{sec:experiments}

\subsection{Main result}\label{sec:main}

\begin{table*}[t]
    \caption{Model performance obtained using EBES. Results are averaged over 20 runs, with the best hyperparameters determined through HPO. Statistically indistinguishable ($p > 0.01$) results share the same superscripts, indicating the method's rank for each dataset. The best-performing methods for each dataset are highlighted. Methods are sorted according to their average rank across all datasets. Note: 4/20 runs of mTAND on the Pendulum dataset were excluded due to non-convergence (<20\% accuracy). Number of learnable parameters presented in \Cref{tab:learnable_parameters}.
}
    \label{tab:main}
    \centering
    \resizebox{\textwidth}{!}{%
\begin{tabular}{r|ccccc|ccc|cc}
\toprule
Category & \multicolumn{5}{c|}{Discrete \ES} & \multicolumn{3}{c|}{Continuous \ES} & \multicolumn{2}{c}{Time Series} \\
\midrule
Dataset & \textbf{MBD} & \textbf{Retail} & \textbf{Age} & \textbf{Taobao} & \textbf{BPI17} & \textbf{PhysioNet2012} & \textbf{MIMIC-III} & \textbf{Pendulum} & \textbf{ArabicDigits} & \textbf{ElectricDevices} \\
\footnotesize{Metric} & \footnotesize{Mean ROC AUC} & \footnotesize{Accuracy} & \footnotesize{Accuracy} & \footnotesize{ROC AUC} & \footnotesize{ROC AUC} & \footnotesize{ROC AUC} & \footnotesize{ROC AUC} & \footnotesize{Accuracy} & \footnotesize{Accuracy} & \footnotesize{Accuracy} \\
\midrule
\textbf{CoLES} & $0.826 \pm 0.001^{2}$ & \cellcolor{lightgray} $\mathbf{0.553 \pm 0.002^{1}}$ & \cellcolor{lightgray} $\mathbf{0.634 \pm 0.005^{1}}$ & \cellcolor{lightgray} $\mathbf{0.713 \pm 0.002^{1}}$ & $0.742 \pm 0.010^{3,4}$ & $0.840 \pm 0.004^{2,3}$ & \cellcolor{lightgray} $\mathbf{0.902 \pm 0.001^{1}}$ & $0.740 \pm 0.013^{2}$ & \cellcolor{lightgray} $\mathbf{0.983 \pm 0.004^{1,2}}$ & \cellcolor{lightgray} $\mathbf{0.729 \pm 0.019^{1,2}}$ \\
\textbf{GRU} & \cellcolor{lightgray} $\mathbf{0.827 \pm 0.001^{1}}$ & $0.543 \pm 0.002^{2}$ & $0.626 \pm 0.004^{2}$ & \cellcolor{lightgray} $\mathbf{0.713 \pm 0.004^{1}}$ & \cellcolor{lightgray} $\mathbf{0.754 \pm 0.004^{1}}$ & \cellcolor{lightgray} $\mathbf{0.846 \pm 0.004^{1}}$ & \cellcolor{lightgray} $\mathbf{0.901 \pm 0.002^{1}}$ & $0.683 \pm 0.031^{3}$ & $0.975 \pm 0.003^{4}$ & \cellcolor{lightgray} $\mathbf{0.741 \pm 0.013^{1}}$ \\
\textbf{MLEM} & $0.824 \pm 0.001^{3}$ & $0.544 \pm 0.002^{2}$ & \cellcolor{lightgray} $\mathbf{0.634 \pm 0.003^{1}}$ & \cellcolor{lightgray} $\mathbf{0.713 \pm 0.004^{1}}$ & \cellcolor{lightgray} $\mathbf{0.753 \pm 0.005^{1,2}}$ & \cellcolor{lightgray} $\mathbf{0.846 \pm 0.007^{1}}$ & $0.899 \pm 0.002^{2}$ & $0.676 \pm 0.017^{3}$ & $0.978 \pm 0.002^{3}$ & \cellcolor{lightgray} $\mathbf{0.736 \pm 0.014^{1}}$ \\
\textbf{Transformer} & $0.821 \pm 0.002^{4}$ & $0.536 \pm 0.006^{3,4}$ & $0.621 \pm 0.006^{2}$ & $0.692 \pm 0.013^{3,4}$ & \cellcolor{lightgray} $\mathbf{0.749 \pm 0.006^{1,2,3}}$ & $0.838 \pm 0.008^{2,3,4}$ & $0.894 \pm 0.002^{3}$ & $0.658 \pm 0.019^{4}$ & \cellcolor{lightgray} $\mathbf{0.986 \pm 0.004^{1,2}}$ & $0.710 \pm 0.024^{2}$ \\
\textbf{Mamba} & $0.820 \pm 0.003^{4}$ & $0.538 \pm 0.003^{3}$ & $0.609 \pm 0.006^{3}$ & $0.693 \pm 0.023^{2,3}$ & $0.737 \pm 0.012^{4,5}$ & $0.835 \pm 0.006^{3,4}$ & $0.895 \pm 0.002^{3}$ & $0.687 \pm 0.017^{3}$ & $0.983 \pm 0.005^{2}$ & $0.716 \pm 0.022^{2}$ \\
\textbf{ConvTran} & $0.816 \pm 0.002^{5}$ & $0.534 \pm 0.005^{4}$ & $0.603 \pm 0.006^{4}$ & $0.703 \pm 0.009^{2}$ & $0.748 \pm 0.006^{2,3}$ & $0.837 \pm 0.006^{2,3,4}$ & $0.892 \pm 0.005^{3,4}$ & $0.674 \pm 0.028^{3,4}$ & \cellcolor{lightgray} $\mathbf{0.986 \pm 0.003^{1}}$ & $0.711 \pm 0.019^{2}$ \\
\textbf{mTAND} & $0.798 \pm 0.002^{7}$ & $0.519 \pm 0.003^{6}$ & $0.582 \pm 0.009^{5}$ & $0.672 \pm 0.010^{5}$ & $0.738 \pm 0.005^{4}$ & $0.841 \pm 0.005^{2}$ & $0.888 \pm 0.003^{4,5}$ & \cellcolor{lightgray} $\mathbf{0.777 \pm 0.031^{1} *}$ & $0.951 \pm 0.010^{5}$ & $0.631 \pm 0.019^{3}$ \\
\textbf{PrimeNet} & $0.780 \pm 0.006^{8}$ & $0.521 \pm 0.003^{6}$ & $0.583 \pm 0.011^{5}$ & $0.681 \pm 0.010^{4}$ & $0.730 \pm 0.006^{5}$ & $0.839 \pm 0.004^{2,3,4}$ & $0.887 \pm 0.004^{5}$ & $0.600 \pm 0.026^{5}$ & $0.958 \pm 0.009^{5}$ & $0.636 \pm 0.016^{3}$ \\
\textbf{MLP} & $0.809 \pm 0.001^{6}$ & $0.526 \pm 0.002^{5}$ & $0.581 \pm 0.007^{5}$ & $0.659 \pm 0.035^{5}$ & $0.737 \pm 0.004^{4}$ & $0.835 \pm 0.004^{4}$ & $0.881 \pm 0.001^{6}$ & $0.186 \pm 0.006^{6}$ & $0.760 \pm 0.011^{6}$ & $0.437 \pm 0.019^{4}$ \\
\bottomrule
\end{tabular}
}
\end{table*}

The results of \textbf{Final Evaluation} from Section~\ref{sec:mod_eval} are presented in Table~\ref{tab:main}, where methods are ranked from top to least performing.
Alongside the mean performance, we report the method's rank as a superscript. 
We performed pairwise Mann--Whitney $U$ test~\citep{mann1947test} with Benjamini--Hochberg correction~\citep{benjamini1995controlling}, methods with no significant performance difference ($p > 0.01$) share the same superscript. 
We make the following observations.

\textbf{GRU-based models dominate \ES classification.} The top three performing methods are all based on GRU with varying pre-training strategies. This is intriguing, as other domains of sequential modeling are often dominated by attention-based or convolution-based architectures. CoLES demonstrates improved metrics on tasks where the target is a characteristic of the observed sequence, such as Age, Pendulum, and Retail. However, on datasets where the target is related to future events, such as MBD, Taobao, BPI17, PhysioNet, and MIMIC-III, pre-training does not provide a significant boost. This is due to CoLES' pre-training procedure, which treats all subsequences of a sequence as belonging to the same class as the full sequence.
Even though MLEM utilizes pre-trained CoLES components, it performs similarly to CoLES and does not outperform both GRU and CoLES simultaneously on any dataset.

\textbf{Transformer and Mamba are next in the ranking,} suggesting that these architectures may be less suited for \ES classification, although they dominate the text domain, which is also sequential data.

\textbf{Success in TS classification doesn't guarantee \ES classification performance.} mTAND~\citep{shukla2021multitime} excels on the Pendulum dataset due to its architecture tailored for modeling the time component, making it well-suited for datasets like Pendulum. Similarly, \mbox{ConvTran} performs best on the Multivariate TS dataset, ArabicDigits, for which it was specifically designed. On this dataset, other models with attention mechanisms, such as Transformer, also perform well. However, both mTAND and ConvTran struggle on other datasets, indicating that these methods may not be as effective for general event sequence classification. This suggests that state-of-the-art models designed for time series analysis do not necessarily perform well on event sequence classification tasks, underscoring the need to treat \ES as a distinct domain.
 
Notably, even though mTAND was explicitly designed for PhysioNet2012 and MIMIC-III, and ConvTran was theoretically better suited for ElectricDevices, both were outperformed by \ES-specific models. This asymmetry suggests an intriguing direction: rather than attempting to adapt TS methods to \ES tasks, it may be worthwhile to explore the reverse—applying \ES models to TS problems.

\textbf{MLP's performance suggests effective \ES classification with aggregated statistics.} The MLP performs relatively well, typically within 5\% of the top-performing method on most real-world datasets. This suggests that \ES classification can be effectively carried out using aggregated statistics along temporal dimensions, a practice commonly employed in industrial applications with boosting models~\citep{abidar2023predicting}. The difference in performance between MLP and mTAND on the Pendulum dataset further supports this idea, since we cannot apply such aggregation approach to this dataset. 

\textbf{The PhysioNet2012 dataset's effectiveness for \ES classification evaluation is questionable.}
Despite its widespread use in modeling irregularly sampled time series, all methods, including the MLP, achieve closely ranked results, suggesting limited ability to distinguish model performance. 
We include this dataset precisely because of its prevalence in the literature, aiming to demonstrate its shortcomings and highlight the need for more discriminative datasets in \ES classification tasks.

\subsection{The Role of Sequence Order and Time in \ES classification}
To understand how important the order of events and their timestamps are in \ES classification, we conducted three experiments: 1) we tested models trained on original sequences with shuffled event orders; 2) we tested models trained on original sequences with random timestamps; and 3) to check whether temporal information matters at all, we removed timestamps and retrained models on shuffled sequences, ensuring the model had no information about temporal structure.

\begin{table*}[t]
\caption{Testing on Permuted Sequences. Models were trained on non-permuted data; only the test set was permuted. We report performance difference relative to metrics obtained on not permuted sequences. Only values with statistically significant difference ($p < 0.01$) in performance are highlighted.}
\label{tab:abl_perm_kl_rel}
\centering
\resizebox{\textwidth}{!}{%

\begin{tabular}{r|ccccc|ccc|cc}
\toprule
Category & \multicolumn{5}{c|}{Discrete \ES} & \multicolumn{3}{c|}{Continuous \ES} & \multicolumn{2}{c}{Time Series} \\
\midrule
Dataset & \textbf{MBD} & \textbf{Retail} & \textbf{Age} & \textbf{Taobao} & \textbf{BPI17} & \textbf{PhysioNet2012} & \textbf{MIMIC-III} & \textbf{Pendulum} & \textbf{ArabicDigits} & \textbf{ElectricDevices} \\
\footnotesize{Metric} & \footnotesize{Mean ROC AUC} & \footnotesize{Accuracy} & \footnotesize{Accuracy} & \footnotesize{ROC AUC} & \footnotesize{ROC AUC} & \footnotesize{ROC AUC} & \footnotesize{ROC AUC} & \footnotesize{Accuracy} & \footnotesize{Accuracy} & \footnotesize{Accuracy} \\
\midrule
\textbf{CoLES} & $-0.09 \%$ & \cellcolor{gray!25}$-1.57 \%$ & \cellcolor{gray!25}$-1.63 \%$ & \cellcolor{gray!15}$-0.49 \%$ & \cellcolor{gray!25}$-4.66 \%$ & \cellcolor{gray!25}$-2.36 \%$ & \cellcolor{gray!25}$-1.86 \%$ & \cellcolor{gray!100}$-84.49 \%$ & \cellcolor{gray!75}$-33.86 \%$ & \cellcolor{gray!100}$-68.79 \%$ \\
\textbf{GRU} & $-0.10 \%$ & \cellcolor{gray!25}$-2.25 \%$ & \cellcolor{gray!25}$-1.15 \%$ & \cellcolor{gray!15}$-0.67 \%$ & \cellcolor{gray!25}$-4.46 \%$ & \cellcolor{gray!25}$-1.49 \%$ & \cellcolor{gray!25}$-4.24 \%$ & \cellcolor{gray!100}$-76.09 \%$ & \cellcolor{gray!75}$-46.88 \%$ & \cellcolor{gray!100}$-69.46 \%$ \\
\textbf{MLEM} & $-0.30 \%$ & \cellcolor{gray!25}$-2.57 \%$ & \cellcolor{gray!25}$-1.52 \%$ & \cellcolor{gray!15}$-0.89 \%$ & \cellcolor{gray!25}$-3.80 \%$ & \cellcolor{gray!25}$-1.71 \%$ & \cellcolor{gray!25}$-1.43 \%$ & \cellcolor{gray!100}$-81.84 \%$ & \cellcolor{gray!75}$-37.81 \%$ & \cellcolor{gray!100}$-65.17 \%$ \\
\textbf{Transformer} & $-0.00 \%$ & $-0.09 \%$ & $-0.00 \%$ & $-0.05 \%$ & $-0.00 \%$ & $0.03 \%$ & $-0.00 \%$ & $-0.00 \%$ & \cellcolor{gray!50}$-15.12 \%$ & \cellcolor{gray!75}$-25.26 \%$ \\
\textbf{Mamba} & $-0.06 \%$ & \cellcolor{gray!25}$-2.44 \%$ & $-1.20 \%$ & $-0.00 \%$ & \cellcolor{gray!50}$-9.56 \%$ & \cellcolor{gray!15}$-0.65 \%$ & \cellcolor{gray!25}$-3.04 \%$ & \cellcolor{gray!100}$-82.14 \%$ & \cellcolor{gray!100}$-53.37 \%$ & \cellcolor{gray!100}$-54.18 \%$ \\
\textbf{ConvTran} & \cellcolor{gray!50}$-7.28 \%$ & \cellcolor{gray!75}$-29.02 \%$ & \cellcolor{gray!50}$-9.55 \%$ & \cellcolor{gray!25}$-4.51 \%$ & \cellcolor{gray!50}$-17.04 \%$ & $-0.47 \%$ & \cellcolor{gray!50}$-8.21 \%$ & \cellcolor{gray!100}$-77.61 \%$ & \cellcolor{gray!100}$-60.45 \%$ & \cellcolor{gray!100}$-68.66 \%$ \\
\textbf{mTAND} & \cellcolor{gray!50}$-5.05 \%$ & \cellcolor{gray!75}$-28.09 \%$ & \cellcolor{gray!50}$-8.95 \%$ & \cellcolor{gray!25}$-4.13 \%$ & \cellcolor{gray!50}$-9.07 \%$ & \cellcolor{gray!25}$-4.13 \%$ & \cellcolor{gray!50}$-5.05 \%$ & \cellcolor{gray!100}$-82.57 \%$ & \cellcolor{gray!100}$-59.12 \%$ & \cellcolor{gray!100}$-56.04 \%$ \\
\textbf{PrimeNet} & \cellcolor{gray!25}$-4.08 \%$ & \cellcolor{gray!75}$-26.41 \%$ & \cellcolor{gray!50}$-7.82 \%$ & \cellcolor{gray!25}$-2.12 \%$ & \cellcolor{gray!25}$-4.73 \%$ & \cellcolor{gray!25}$-3.95 \%$ & \cellcolor{gray!25}$-3.72 \%$ & \cellcolor{gray!100}$-75.88 \%$ & \cellcolor{gray!100}$-53.38 \%$ & \cellcolor{gray!100}$-54.38 \%$ \\
\textbf{MLP} & $-0.00 \%$ & $-0.00 \%$ & $-0.00 \%$ & $-0.00 \%$ & $-0.00 \%$ & $-0.00 \%$ & $-0.00 \%$ & $-0.00 \%$ & $-0.00 \%$ & $-0.00 \%$ \\
\bottomrule
\end{tabular}

}
\end{table*}

\begin{table*}[t]
    \caption{Testing on Random Timestamps. Models were trained on original data; only the test set has random timestamps. We report performance difference relative to metrics obtained on original sequences. Only values with statistically significant difference ($p < 0.01$) in performance are highlighted.}
    \label{tab:abl_time}
    \centering
    \resizebox{\textwidth}{!}{%
\begin{tabular}{r|ccccc|ccc|cc}
\toprule
Category & \multicolumn{5}{c|}{Discrete \ES} & \multicolumn{3}{c|}{Continuous \ES} & \multicolumn{2}{c}{Time Series} \\
\midrule
Dataset & \textbf{MBD} & \textbf{Retail} & \textbf{Age} & \textbf{Taobao} & \textbf{BPI17} & \textbf{PhysioNet2012} & \textbf{MIMIC-III} & \textbf{Pendulum} & \textbf{ArabicDigits} & \textbf{ElectricDevices} \\
\footnotesize{Metric} & \footnotesize{Mean ROC AUC} & \footnotesize{Accuracy} & \footnotesize{Accuracy} & \footnotesize{ROC AUC} & \footnotesize{ROC AUC} & \footnotesize{ROC AUC} & \footnotesize{ROC AUC} & \footnotesize{Accuracy} & \footnotesize{Accuracy} & \footnotesize{Accuracy} \\
\midrule
\textbf{PrimeNet} & $-0.72 \%$ & $-0.07 \%$ & $-0.12 \%$ & $-0.15 \%$ & $-0.30 \%$ & $0.09 \%$ & $-0.40 \%$ & \cellcolor{gray!100}$-66.34 \%$ & \cellcolor{gray!100}$-28.86 \%$ & \cellcolor{gray!75}$-5.62 \%$ \\
\textbf{mTAND} & \cellcolor{gray!25}$-0.45 \%$ & $-0.01 \%$ & $-0.06 \%$ & $-0.91 \%$ & $-0.00 \%$ & $-0.08 \%$ & $-0.23 \%$ & \cellcolor{gray!100}$-56.79 \%$ & \cellcolor{gray!75}$-7.44 \%$ & \cellcolor{gray!75}$-6.11 \%$ \\
\bottomrule
\end{tabular}
}
\end{table*}

\begin{table*}[t]
\centering
\caption{Training on Permuted Sequences without Timestamps. The GRU model with the best hyperparameters had the time feature removed and was then trained from scratch in two settings: \textit{with} and \textit{without} permuting both the training and test sequences. We report performance difference relative to metrics obtained on original sequences. Only values with statistically significant difference ($p < 0.01$) in performance are highlighted.}
\label{tab:abl_perm_gru}
\resizebox{\textwidth}{!}{%
\begin{tabular}{r|ccccc|ccc|cc}
\toprule
Category & \multicolumn{5}{c|}{Discrete \ES} & \multicolumn{3}{c|}{Continuous \ES} & \multicolumn{2}{c}{Time Series} \\
\midrule
Dataset & \textbf{MBD} & \textbf{Retail} & \textbf{Age} & \textbf{Taobao} & \textbf{BPI17} & \textbf{PhysioNet2012} & \textbf{MIMIC-III} & \textbf{Pendulum} & \textbf{ArabicDigits} & \textbf{ElectricDevices} \\
\footnotesize{Metric} & \footnotesize{Mean ROC AUC} & \footnotesize{Accuracy} & \footnotesize{Accuracy} & \footnotesize{ROC AUC} & \footnotesize{ROC AUC} & \footnotesize{ROC AUC} & \footnotesize{ROC AUC} & \footnotesize{Accuracy} & \footnotesize{Accuracy} & \footnotesize{Accuracy} \\
\midrule
\textbf{GRU w/o time} & \cellcolor{gray!25}$-0.89 \%$ & $-0.00 \%$ & \cellcolor{gray!15}$-0.44 \%$ & \cellcolor{gray!50}$-3.85 \%$ & $-0.00 \%$ & $-0.00 \%$ & \cellcolor{gray!15}$-0.27 \%$ & \cellcolor{gray!100}$-59.43 \%$ & $0.04 \%$ & $-0.00 \%$ \\
\textbf{GRU w/o time w/ perm.} & \cellcolor{gray!25}$-0.96 \%$ & $0.50 \%$ & $0.62 \%$ & \cellcolor{gray!50}$-1.54 \%$ & $-0.45 \%$ & $-0.22 \%$ & \cellcolor{gray!50}$-1.25 \%$ & \cellcolor{gray!100}$-63.87 \%$ & \cellcolor{gray!50}$-1.28 \%$ & \cellcolor{gray!75}$-16.00 \%$ \\
\bottomrule
\end{tabular}
}
\end{table*}

\subsubsection{Testing on Permuted Sequences} 
We evaluated pre-trained models on perturbed sequences \emph{without fine-tuning}.
Missing values were filled prior to shuffling, and time was added as a numerical feature before shuffling.
For all runs, the last events were kept in their original positions, as some models use the last hidden state in the aggregation step.
Results are presented in Table~\ref{tab:abl_perm_kl_rel}. 
We made two key observations:

\begin{itemize}[leftmargin=*, itemsep=0pt, parsep=0pt]
\item The datasets can be clearly divided into two groups based on the performance drop.
The group with the largest drop includes time series datasets (ArabicDigits and ElectricDevices) and a synthetic dataset designed to evaluate the ability to capture temporal structure (Pendulum). The second group consists of real-world \ES data, where the performance drop is relatively small, indicating that sequence order is not as critical for \ES classification as it might seem to be.
\item Models specifically designed for time series exhibit a larger drop in performance across all datasets compared to \ES models.
\end{itemize}
This indicates that capturing the properties of \ES requires specific models, further supporting the importance of \ES as a distinct domain.

The Transformer model experienced a minimal drop on \ES datasets due to its attention mechanism. Notably, the drop is zero on datasets where positional embedding (PE) was not used (the use of PE was determined during HPO).
The MLP model did not experience any performance drop because sequence order is inherently unimportant for aggregation.
Notably, the MBD dataset did not experience a significant drop with most methods.

\subsubsection{Testing on Random Timestamps.}
Two methods are specifically designed to model the time component in our selection of methods: mTAND~\citep{shukla2021multitime} and PrimeNet~\citep{chowdhury2023primenet}.
We evaluated them on test data with noisy timestamps, where the original timestamps were replaced with random values sorted in ascending order. The results are presented in Table~\ref{tab:abl_time}. While time is important for these models on the ArabicDigits, ElectricDevices and synthetic Pendulum dataset, it did not contribute significantly to the other datasets.

We conclude that methods specifically designed to work with time do not effectively capture temporal dependencies on real-world \ES datasets. This emphasizes the importance of developing or testing new methods on \ES that can model the time component on real-world datasets.

\subsubsection{Training on Permuted Sequences without Timestamps} The second experiment further analyzed datasets to determine if order information is important or if sequences can be treated as a \textbf{``bag of words''.}
We selected the GRU with the best hyperparameters for each dataset and removed time completely from the model inputs.
The model was then trained from scratch under two conditions: \textbf{with} and \textbf{without} permuting both the training and test sequences.
The results are presented in Table~\ref{tab:abl_perm_gru}.

We observed that for some real-world datasets, the performance drop was not statistically significant.
We speculate that such permutation could even serve as a form of data augmentation, since in some cases mean metrics increased with permutation. This phenomenon was observed in datasets such as Retail, Age, BPI17, and Physionet2012.

On the other hand, datasets like MBD and Taobao showed performance declines when the time feature was removed. However, permuting the sequences on top of that did not cause a significant further drop. This suggests that for these datasets, the time feature is more important than the sequence order itself.

Finally, datasets such as MIMIC-III, Pendulum, ArabicDigits, and ElectricDevices demonstrated a clear performance drop, underscoring the importance of sequence order for the GRU model on those datasets.

From both experiments, we conclude that sequence order is important for \ES classification, but it is less critical than expected for real-world datasets and varies from dataset to dataset.



\subsection{HPO analysis}

\subsubsection{Assessing Architecture Design Choices}
\label{sec:arch_analysis} 

This section evaluates the impact of key architecture design choices—aggregation, normalization, learning rates, and time features—across various models and datasets, providing insights from our hyperparameter optimization (HPO) procedure.

First, we examine the effect of temporal aggregation methods, comparing the mean of all hidden states to the last hidden state. Results show that the choice of aggregation significantly impacts performance. CoLES and MLEM consistently benefit from using the last hidden state across all datasets, as do Physionet2012 and MIMIC-III. In contrast, Mamba performs better with the mean hidden state on all datasets except Physionet2012, while other models exhibit individual preferences. Detailed results are provided in Table~\ref{tab:aggregation}.

Second, we analyze the impact of batch normalization on numerical features. Results indicate that batch normalization improves performance for nearly all methods and datasets, with the exception of Pendulum and ArabicDigits. See Table~\ref{tab:normalization} for further details.

We also investigate the role of time features. Table~\ref{tab:time_to_none} demonstrates that incorporating time significantly enhances performance for MBD, Age, Taobao, MIMIC-III, and Pendulum datasets. However, methods such as Retail, BPI17, and ArabicDigits show no notable benefit. While these results highlight dataset-specific trends, they do not preclude the potential importance of time in other contexts, as alternative integration methods—unexplored in this study—may yield further improvements.

Finally, HPO evaluations consistently identify the learning rate as one of the most important hyperparameter across all runs, as summarized in Table~\ref{tab:LR}. This underscores its critical role in optimizing model performance.

\subsubsection{Dataset Analysis}\label{sec:dataset_analysis}

In this section, we analyze datasets based on data from the \textbf{HPO} and \textbf{Final Evaluation} phases, exploring relationships between metrics from different data subsets. See ~\Cref{appendix: hpo_corr}.

During the HPO step, we observe overfitting for most datasets, as \texttt{train} metrics increase while \texttt{train-val} metrics plateau, as seen in Figure~\ref{fig:physionet2012_hpo_corr} on the left. This supports the use of early stopping.

Metrics of \texttt{hpo-val} and \texttt{test} subsets (third column in Figure~\ref{fig:physionet2012_hpo_corr}) are strongly correlated unless the test set is sampled out-of-time, as seen for the Taobao and BPI17 dataset. Here, \texttt{hpo-val} and \texttt{test} metrics lack a clear linear trend, but \texttt{train-val} and \texttt{hpo-val} metrics do, suggesting a distribution shift in the test set.

For most datasets, in the \textbf{Final Evaluation} phase (fourth column in Figure~\ref{fig:physionet2012_hpo_corr}), validation and test set metrics exhibit a linear trend, except for PhysioNet2012, where different validation metrics attribute to similar test metrics. This supports our observations in Section~\ref{sec:main}, where results for most models are not statistically distinguishable for most methods on PhysioNet2012.

\subsection{Data Scaling Results}

\begin{figure}[t]
    \centering
    \begin{subfigure}[b]{0.4\textwidth}
        \centering
        \includegraphics[width=\textwidth]{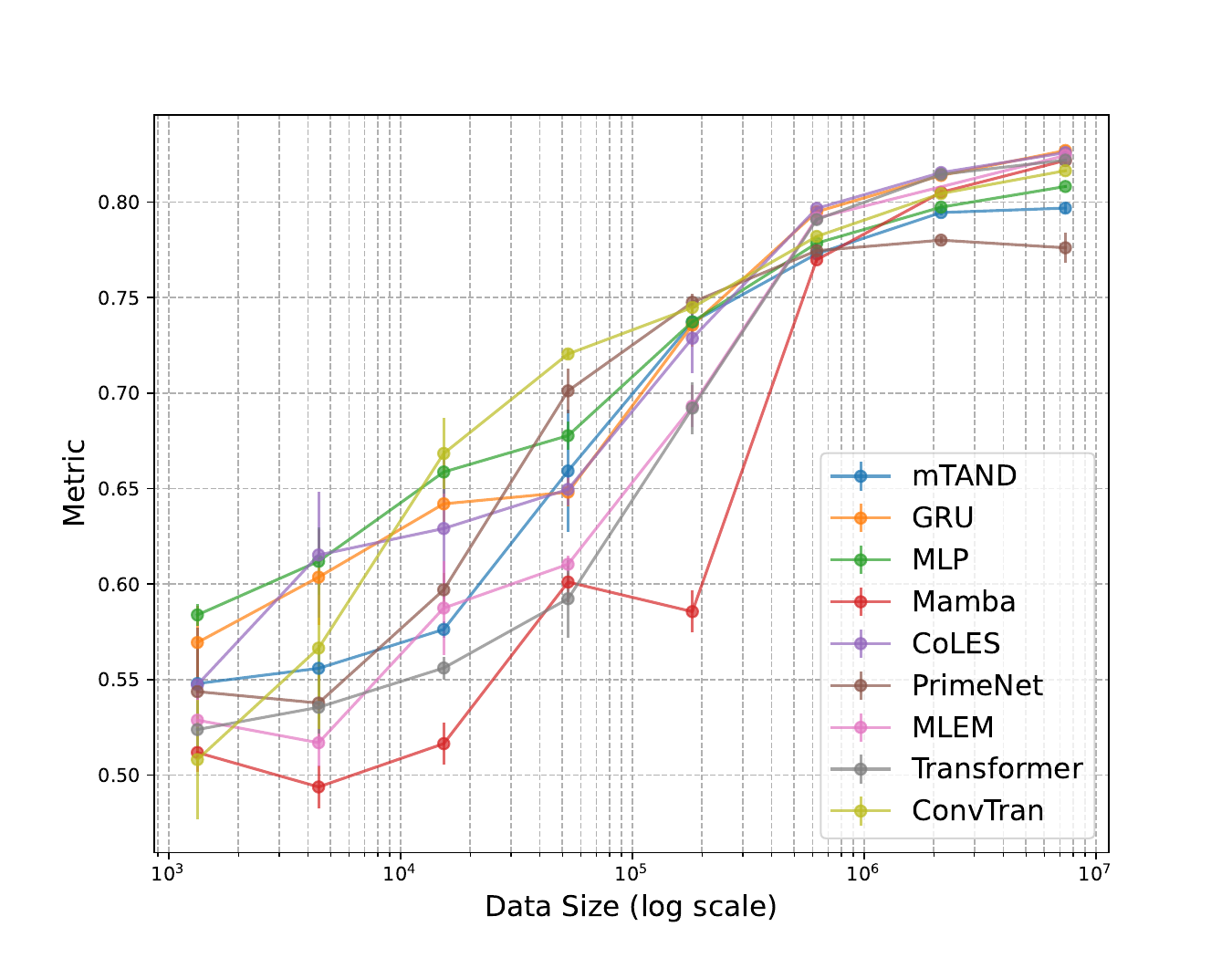}
        \caption{MBD dataset}
        \label{fig:mbd}
    \end{subfigure}
    \begin{subfigure}[b]{0.4\textwidth}
        \centering
        \includegraphics[width=\textwidth]{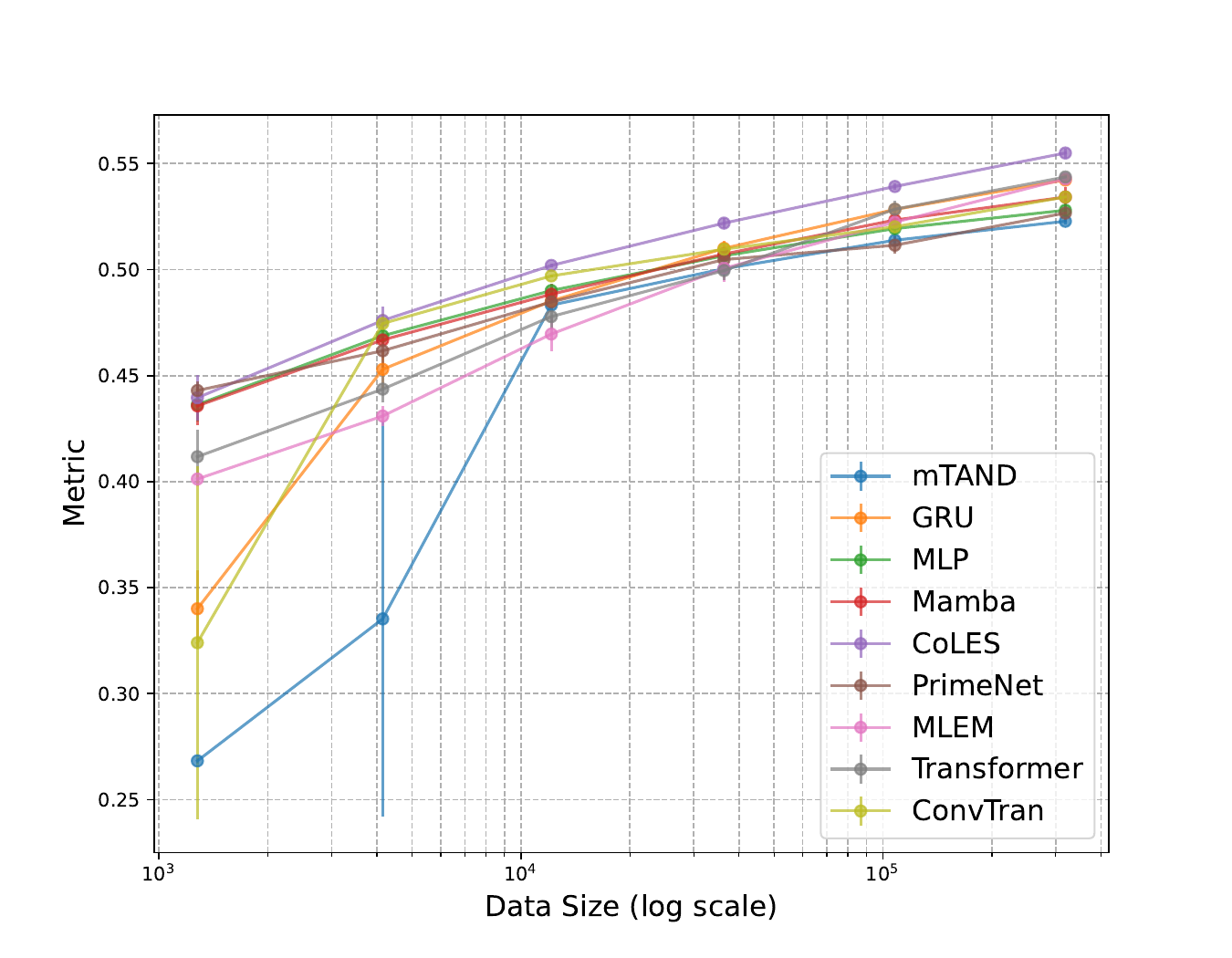}
        \caption{Retail dataset}
        \label{fig:retail}
    \end{subfigure}
    \caption{Data Scaling Results. We take models with the best hyperparameters and retrain them on subsets of varying sizes. The number of sequences is presented on a log scale. Standard deviation across 3 runs is indicated by vertical lines.}
    \label{fig:scaling}
\end{figure}

To study the scaling properties of various models, we evaluated each model trained with different numbers of sequences. We focused on two biggest real-world dataset in our benchmark: Retail and MBD. We sampled different subsets, each containing progressively more data. Each model was trained from scratch on different-sized subsets with Monte Carlo cross-validation using three random seeds. 

A common approach is to estimate model performance with a fixed data size. However, as seen in Figure~\ref{fig:scaling}, while all models improve with the growth of the data, their ranking does not stay the same, except for CoLES on the Retail Dataset, where it consistently demonstrates superior performance. With some data size, even MLP becomes a top performer. Most models, except for MLP, mTAND, and PrimeNet, converge to similar performance on the MBD dataset given a large data size.
It is worth noting that for each dataset, we used the best hyperparameters found for each model when the dataset was at its full size.

The standard deviation, decreases as the data size increases and models perform very differently with smaller subsets. This makes evaluating model performance on relatively small datasets more prone to misleading results.


\section{Related work}\label{sec:related_work}

The UCR Time Series Archive~\citep{UCR}, a widely used benchmark for time series classification, comprises 128 datasets designed to evaluate algorithms on univariate time series data. 
While this archive has been instrumental in advancing time series research, it does not address the unique challenges posed by \ES data, such as irregular intervals and heterogeneous feature types.
Similarly, the \texttt{torchtime} package~\citep{darke2022benchmark} extends the utility of UEA and UCR datasets by providing reproducible implementations for PyTorch.
However, it remains focused on the classification of time series and does not cover the complexity of \ES.

The temporal point process (TPP) formalism is often used to model various types of \ES. 
EasyTPP~\citep{xue2024easytpp} is a recent benchmark targeting streams of discrete events, offering a centralized repository for evaluating TPP models.
However, EasyTPP focuses primarily on next-event prediction tasks, such as forecasting the time and type of the next event, rather than sequence-level classification. 
Many datasets in EasyTPP consist of only one categorical feature per event and a timestamp.
Moreover, the majority of them lack established targets attributed to entire sequences, thus making them unsuitable for evaluating \ES classification methods.

Another closely related benchmark is HoTPP~\citep{karpukhin2024hotpp}, which evaluates TPP models on long-term prediction tasks rather than next-event prediction.
While HoTPP addresses a different aspect of event sequence modeling, it still does not focus on sequence-level classification, which is our primary goal.

\section{Limitations and future work}
We acknowledge that conducting a full hyperparameter optimization process requires substantial computational resources, which may not be available to all users. Additionally, using a random subset of the full training set as a proxy for the test set is not always the best method for estimating hyperparameters, as seen in \Cref{fig:taobao_hpo_corr}. The development of more efficient strategies for proper model evaluation and more robust procedures to address potential data shifts could be a promising direction for future research.

Our work focuses solely on one task—\ES classification while there are various tasks applied to \ES. We leave this for future work.

\section{Conclusion}

In this work, we introduced \textbf{EBES}, an open and comprehensive benchmark designed to enable standardized and transparent comparisons of models for \ES classification. By curating a diverse range of datasets and models, EBES provides a robust framework for evaluating various approaches' performance while also offering a user-friendly interface and a rich library for easy integration of new datasets and methods.

First, EBES establishes \ES classification as a distinct domain requiring specialized methods. Our analysis demonstrates that \ES has unique properties and behaviors that set it apart from related sequential data types like time series. This is supported by statistically proven results showing that models perform differently on \ES compared to other sequential data. In particular, GRU-based models dominate \ES classification, while time series-specific models perform worse. We also demonstrated that the importance of time and sequential structure varies across real-world datasets and tends to be lower than expected on \ES datasets compared to time series datasets.
This variability underscores the necessity of developing or adapting models that inherently account for the time component in real-world scenarios rather than relying on methods initially designed for time series or discrete event streams.

We demonstrated how HPO analysis can reveal hidden trends in architecture design and datasets structure. 
For instance, results on the PhysioNet2012 dataset reveal a limited ability to distinguish model performance, suggesting that it may not be suitable for benchmarking \ES-specific methods. Additionally, the observed distribution shifts in out-of-time splits for datasets like Taobao and BPI17 emphasize the importance of accounting for real-world complexities during model validation and hyperparameter optimization. 

The insights gained from EBES provide a foundation for future research, encouraging the development of novel architectures and techniques that address the unique challenges posed by event sequences. 
As researchers continue to explore this domain, we hope that EBES will catalyze innovation and guide the community toward more effective solutions for understanding and leveraging event sequence data.

\bibliographystyle{ACM-Reference-Format}
\bibliography{biblio}

\appendix
\section{Models description}\label{sec:app_models}

\paragraph{GRU}
We have chosen to use the GRU as one of our base models due to its proven effectiveness in encoding time-ordered sequences~\cite{babaev2022coles, rubanova2019latent,tonekaboni2021unsupervised,NEURIPS2019_c9efe5f2,udovichenko2024seqnas}. 
In recent study on neural architecture search~\cite{udovichenko2024seqnas}), authors demonstrated that architectures with RNN blocks tend to exhibit higher performance on average on EvS assesment task.

\paragraph{CoLES}
The contrastive pretraining method for sequential data was proposed by~\cite{babaev2022coles}.
We specifically focus on this method due to its superior performance compared to other contrastive approaches demonstrated in the work.
CoLES learns to encode a sequence into a latent vector by bringing sub-sequences of the same sequence closer in the embedding space while pushing sub-sequences from different sequences further apart. 

\paragraph{MLEM}
The Multimodal Learning Event Model~\cite{moskvoretskii2024selfsupervised} is a recently proposed method for Event Sequences that unifies contrastive learning with generative modeling. It treats generative pre-training and contrastive learning as distinct modalities. First, a contrastive encoder is trained, followed by an encoder-decoder that learns latent states using reconstruction loss while aligning with contrastive embeddings to enhance the embedding information.

\paragraph{Transformer}
As another base model, we added the transformer architecture \cite{vaswani2017attention}, as attention-based architectures are one of the most common models in sequential modeling. We took the basic PyTorch implementation and provided three options for positional encoding (chosen in HPO): sinusoidal positional encoding added to features, sinusoidal positional encoding concatenated along the temporal dimension, learned positional embeddings, and no positional encoding at all.

\paragraph{MLP}
The models applied 3 linear layers with the ReLU nonlinearity and dropouts in between to the aggregated embeddings obtained right after the preprocessing block.
So effectively the model is just a basic MLP applied to aggregations.
Models for \ES handles the sequential nature of data in a special way, ofthen considering the exact time intervals between the events, so we were interested in the performance of the model, that consciously discards the sequential nature of data.

\paragraph{Mamba}
Mamba~\cite{gu2023mamba} is a recent state-space model (SSM) that has been designed for efficient handling of complex, long sequences.
It incorporates selective state spaces to deliver top-notch performance across different modalities, including language, audio, and genomics, outperforming Transformers in some scenarios.
For the best of our knowledge Mamba has not been applied to EvS classification previously, however, we believe that type of models worth of investigating. 

\paragraph{mTAND}  
Authors in~\cite{shukla2021multitime} proposed an architecture which learns an embedding of continuous-time values and utilizes an attention mechanism to produce a fixed-length representation of a time series.
This procedure is specifically designed to deal with ISTS and has been shown to outperform numerous ordinary differential equations-based models such as Latent ODE and ODE-RNN~\cite{rubanova2019latent}.

\paragraph{PrimeNet}
The method proposed in~\cite{chowdhury2023primenet} also, falls under the category of self-supervised.
It utilizes time-sensitive contrastive pretraining and enhances pretraining procedure with data reconstruction tasks to facilitate the usage of unlabeled data.
Authors modify mTAN architecture by replacing an RNN block with Feature-Feature Attention.

\paragraph{ConvTran}
We have chosen to use the ConvTran \cite{foumani2024improving}  due to its state-of-the-art performance in multivariate time series classification (MTSC) tasks. ConvTran combines novel position encoding techniques, specifically time Absolute Position Encoding (tAPE) and efficient Relative Position Encoding (eRPE), with convolution-based input encoding. This combination enhances the model's ability to capture both the temporal ordering and the data embedding of time series data effectively. 
\section{Datasets Description}\label{sec:app_data}

\paragraph{PhysioNet2012} dataset\footnote{\url{https://physionet.org/content/challenge-2012/1.0.0/}} was first intruduced in~\cite{goldberger2000physiobank}.
It includes multivariate time series data with 37 variables gathered from intensive-care unit (ICU) records. 
Each record contains measurements taken at irregular intervals during the first 48 hours of ICU admission.
We used \texttt{set-a} as a train set and \texttt{set-b} as a test set.
Both sets contain 4000 labeled sequences.

\paragraph{MIMIC-III} dataset\footnote{\url{https://physionet.org/content/mimiciii/1.4/}}~\cite{johnson2016mimic} consists of multivariate time series data featuring sparse and irregularly sampled physiological signals, collected at Beth Israel Deaconess Medical Center from 2001 to 2012.
While we aimed to follow the general pipeline outlined in \cite{shukla2018interpolation}, we made several modifications to enhance the accuracy and reproducibility of our approach.
Importantly, we did not alter the original problem statement: we excluded series that last less than 48 hours and used the first 48 hours of observations from the remaining series to predict in-hospital mortality.
These adjustments were necessary to address certain issues and improve the overall robustness of our analysis.

\paragraph{Age} dataset\footnote{\url{https://ods.ai/competitions/sberbank-sirius-lesson}} consists of 44M anonymized credit card transactions representing 50K individuals.
The target is to predict the age group of a cardholder that made the transactions.
The multiclass target label is known only for 30K records, and within this subset the labels are balanced.
Each transaction includes the date, type, and amount being charged.
The dataset was first introduced in scientific literature in work~\cite{babaev2022coles}.

\paragraph{Retail} dataset\footnote{\url{https://ods.ai/competitions/x5-retailhero-uplift-modeling}} comprises 45.8M retail purchases from 400K clients, with the aim of predicting a client's age group based on their purchase history.
Each purchase record includes details such as time, item category, the cose, and loyalty program points received.
The age group information is available for all clients, and the distribution of these groups is balanced across the dataset.
The dataset was first introduced in scientific literature in work~\cite{babaev2022coles}.

\paragraph{MBD} is a multimodal banking dataset introduced in~\cite{mbd}. The dataset contains an industrial-scale number of sequences, with data from more than 1.5 million clients. Each client corresponds to a sequence of events. This multi-modal dataset includes card transactions, geo-position events, and embeddings of dialogs with technical support. The goal is to predict the purchases of four banking products in each month, given the historical data from the previous month. For our analysis, we use only card transactions.

Since we focused on the event sequence classification task, we restricted our setup as follows. To predict the purchases, we use transactions from the preceding month. For example, we use a sequence from June to predict a label by the last day of July. We did not use out-of-time validation, as the labeled time span of the data is less than a year. The authors of the dataset split the data into 5 folds (0--4), we use fold 4 as the test fold.

\paragraph{Taobao} Dataset comprises user behaviors from Taobao, including clicks, purchases, adding items to the shopping cart, and favoriting items. These events were collected between November 18 and December 15. For our analysis, we treat each week of clicks as a sequence and aim to predict payments for the subsequent 7 days following the selected week. The training set encompasses data from November 18 to December 1, while the test set includes clicks from December 2 to December 15.

\paragraph{BPI17} 
We took the dataset from the Business Process Intelligence 2017 Challenge. 
The dataset describes the logs of events related to the business process of negotiating a loan with the customer.
The target is to predict whether the loan offer will be accepted by the customer.
For this dataset, we did an out-of-time test split.

\paragraph{ArabicDigits}
This data set is taken from the UCI repository \cite{bagnall2018uea}. It is derived from sound. Dataset from 8800 (10 digits x 10 repetitions x 88 speakers) time series of 13 Frequency Cepstral Coefficients (MFCCs) had taken from 44 males and 44 females Arabic native speakers between the ages 18 and 40 to represent ten spoken Arabic digit. Each line on the data base represents 13 MFCCs coefficients in the increasing order separated by spaces. This corresponds to one analysis frame.

\paragraph{ElectricDevices}
These problems were taken from data recorded as part of government sponsored study called Powering the Nation. The intention was to collect behavioural data about how consumers use electricity within the home to help reduce the UK's carbon footprint. The data contains readings from 251 households, sampled in two-minute intervals over a month. The data required considerable preprocessing to get into a usable format. We create two distinct types of problem: problems with similar usage patterns (Refrigeration, Computers, Screen) and problems with dissimilar usage patterns (Small Kitchen and Large Kitchen). The aim is that problems with dissimilar usage patterns should be well suited to
time-domain classification, whilst those with similar consumption patterns should be much harder.

\paragraph{Pendulum} \label{sec:pendulum}
Inspired by ~\cite{moskvoretskii2024selfsupervised} we created a pendulum dataset to evaluate time-dependent models. 
The Pendulum dataset is specifically designed for event sequence classification tasks, featuring irregular timestamps and missing values. Its task requires models to consider multiple events for predictions, making it effective in evaluating temporal modelling capabilities.

The dataset simulates damped pendulum motion with varying lengths. Observation times are sampled irregularly using a Hawkes process, emphasizing the importance of accurate event timing for real-world applications. Each sequence in the dataset consists of events represented by time and two normalized coordinates (x, y), with some values randomly dropped. The goal is to predict the damping factor.
We publish the reproducible code to generate the dataset.

To model the Hawkes process, we consider the following intensity function $\lambda(t)$ that is given by~(\ref{eq:hawkes}).
\begin{equation}
    \lambda(t) = \mu + \sum_{t_i < t} \alpha e^{-\beta (t - t_i)}
    \label{eq:hawkes}
\end{equation}

We used following parameters for the Hawkes process: 
\begin{itemize}
    \item $\mu$ is the base intensity;
    \item $\alpha$ is the excitation factor, was chosen to be 0.5;
    \item $\beta$ is the decay factor, was set to 1.
    \item $t_i$ are the times of previous events before time $t$.
\end{itemize}

The time points are sampled within the interval \([0, \text{end time}]\), where the end time is sampled from a uniform distribution $U(3, 5)$. To maintain an approximately constant number of points (30) per sequence, we adjust the base intensity \(\mu\) as follows:

\[
\mu = 30 \times \frac{1 - \alpha}{\text{end time} - 1}
\]

This ensures each sequence has a dynamic time interval but approximately the same number of points, preventing the model from learning the timestamp distribution without using timestamp data.

\medskip

To model the pendulum we consider the second-order differential equation:
\begin{equation}
    \theta'' + \left( \frac{b}{m} \right) \theta' + \left( \frac{g}{L} \right) \sin(\theta) = 0
\end{equation}
where,
\begin{itemize}
    \item $\theta''$ is the Angular Acceleration,
    \item $\theta'$ is the Angular Velocity,
    \item $\theta$ is the Angular Displacement,
    \item $b$ is the Damping Factor,
    \item $g = 9.81 \, \text{m/s}^2$ is the acceleration due to gravity,
    \item $L$ is the Length of pendulum,
    \item $m$ is the Mass of bob in kg.
\end{itemize}

To convert this second-order differential equation into two first-order differential equations, we let $\theta_1 = \theta$ and $\theta_2 = \theta'$, which gives us:
\begin{equation}
    \theta_2' = \theta'' = -\left( \frac{b}{m} \right) \theta_2 - \left( \frac{g}{L} \right) \sin(\theta_1)
\end{equation}
\begin{equation}
    \theta_1' = \theta_2
\end{equation}

Thus, the first-order differential equations for the pendulum simulation are:
\begin{align}
    \theta_2' &= -\left( \frac{b}{m} \right) \theta_2 - \left( \frac{g}{L} \right) \sin(\theta_1) \\
    \theta_1' &= \theta_2
\end{align}

\smallskip

In our simulations, the damping factor \( b \) is sampled from a uniform distribution \( U(1, 3) \), and the mass of the bob \( m = 1 \). The length \( L \) of the pendulum is taken from a uniform distribution \( U(0.5, 10) \), representing a range of possible lengths from 0.5 to 10 meters. The initial angular displacement \( \theta \) is sampled from a uniform distribution \( U(0, 2\pi) \), and the initial angular velocity \( \theta' \) is sampled from a uniform distribution \( U(-\pi, \pi) \), providing a range of initial conditions in radians and radians per second, respectively.

\smallskip

\begin{figure}[t]
  \centering
  \includegraphics[width=0.5\textwidth]{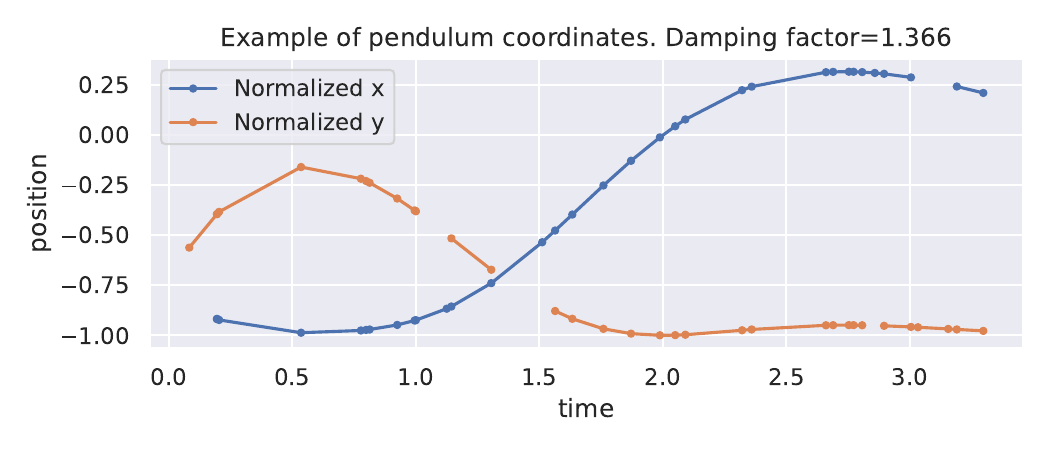}
  \caption{Pendulum motion at various instances, with time steps determined by a Hawkes process.}
  \label{fig:pendulum}
\end{figure}

Our primary objective is to predict the damping factor $b$, using the normalized coordinates $x$ and $y$ on the plane. These coordinates are scaled with respect to the pendulum's length, such that the trajectory of the pendulum is represented in a unitless fashion. This normalization allows us to abstract the pendulum's motion from its actual physical dimensions and instead focus on the pattern of movement. Additionally, we randomly drop 10\% of values for both coordinates. An illustrative example of this motion is presented in Figure~\ref{fig:pendulum}.

\section{HPO details}
\label{appendix:hpo}

Hyperparameter Optimization (HPO) is a critical step in the development and evaluation of machine learning models. It involves systematically searching for the optimal set of hyperparameters that maximize model performance. In this section, we outline our main evaluation methodology and HPO process, which is detailed in Algorithm~\ref{alg:main}.

Our approach includes two main steps: the HPO step and the final evaluation step. In the HPO step, we use the Tree-structured Parzen Estimator (TPE) to efficiently search the hyperparameter space. We split the training dataset into three subsets: \texttt{train} (70\%), \texttt{train-val} (15\%), and \texttt{hpo-val} (15\%). The model is trained on the \texttt{train} set, and its performance is evaluated on the \texttt{train-val} set to determine when to stop training. The \texttt{hpo-val} set is used to update the TPE sampler and guide the selection of hyperparameters.

After the HPO step, we proceed to the final evaluation step. Here, we use the best hyperparameters (BHP) identified in the HPO step to train and evaluate the model multiple times with different random seeds. This ensures that our results are robust and not dependent on a particular random initialization. The training dataset is split into \texttt{train} (85\%) and \texttt{train-val} (15\%) sets, and the model is trained until performance on the \texttt{train-val} set stops improving or until the training budget is exhausted. Finally, we evaluate the model on the test set and report the mean and standard deviation of the test metrics.

For more details about the HPO process, we refer to our Algorithm~\ref{alg:main}.

\begin{algorithm*}
	\caption{Our main evaluation methodolgy and HPO, here $N_{hpo}$ - is HPO budget, $MaxIters$ - training budget, $N_{seeds}$ - a number of iterations for random seed runs.} 
        \label{alg:search}
	\begin{algorithmic}[1]
    \State $MaxIters = 10^5$
    \State $N_{seeds} = 20$
 \State start \textbf{HPO step}
	\State split train dataset randomly into three subsets \texttt{train} (70\%), \texttt{train-val} (15\%) and \texttt{hpo-val} (15\%)
    \State initalize TPE
    \For {$i=1,2,\ldots, N_{hpo}$}
        \State set model hyper parameters with TPE
        \State train a model until performance on \texttt{train-val} set stops improving or until we run out from the budget $MaxIters$.
        \State update TPE sampler using metrics obtained on \texttt{hpo-val} 
		\EndFor
    \State select best hyper parameters \textbf{(BHP)} according to \texttt{hpo-val} metrics
    \State Start \textbf{Final evaluation} step
    \For {$seed=1,2,\ldots, N_{seeds}$}
        \State set a new random $seed$
        \State randomly split train dataset into \texttt{train} (85\%) and \texttt{train-val} (15\%) sets
        \State train a model with \textbf{BHP} until performance on \texttt{train-val} set stops improving or until we run out from the budget $MaxIters$.
        \State evaluate the model on test set
    \EndFor 
    \State Report $mean$ and $std$ of test metrics from \textbf{Final evaluation} step
	\end{algorithmic} 
 \label{alg:main}
\end{algorithm*}




\section{Number of learnable parameters} \label{appendix: n_params}
We report number of learnable parameters of each model from Table~\ref{tab:main} in Table\ref{tab:learnable_parameters}

\begin{table*}[h]
\centering
\caption{Number of learnable parameters for each model from \Cref{tab:main}. Based on successful HPO runs, we report the minimum and maximum number of parameters explored in the search space, along with the number of parameters in the best-performing model. The results indicate that most models do not select the heaviest settings, likely because excessively increasing model size leads to overfitting on the \ES classification task.}
\resizebox{\textwidth}{!}{
\begin{tabular}{rrrrrrrrrrr}
\toprule
 &  & CoLES & GRU & MLEM & Transformer & Mamba & ConvTran & mTAND & PrimeNet & MLP \\
Method & Params &  &  &  &  &  &  &  &  &  \\
\midrule
\multirow[t]{3}{*}{\textbf{MBD}} & \textbf{min} & 2.1e+05 & 3.2e+05 & 9.0e+06 & 1.9e+06 & 4.6e+05 & 1.5e+05 & 3.8e+05 & 3.5e+05 & 4.6e+05 \\
\textbf{} & \textbf{best} & \cellcolor{lightgray}9.1e+06 & \cellcolor{lightgray}1.1e+07 & \cellcolor{lightgray}1.4e+07 & \cellcolor{lightgray}1.4e+08 & \cellcolor{lightgray}5.2e+07 & \cellcolor{lightgray}3.8e+06 & \cellcolor{lightgray}5.1e+06 & \cellcolor{lightgray}1.7e+06 & \cellcolor{lightgray}6.4e+06 \\
\textbf{} & \textbf{max} & 2.9e+07 & 3.0e+07 & 5.1e+07 & 3.0e+08 & 5.7e+07 & 1.7e+07 & 1.2e+07 & 4.2e+06 & 7.9e+06 \\
\cline{1-11}
\multirow[t]{3}{*}{\textbf{Retail}} & \textbf{min} & 1.2e+05 & 8.8e+04 & 1.6e+06 & 2.5e+06 & 2.1e+05 & 8.1e+04 & 1.4e+04 & 3.7e+04 & 2.3e+04 \\
\textbf{} & \textbf{best} & \cellcolor{lightgray}2.3e+06 & \cellcolor{lightgray}3.8e+06 & \cellcolor{lightgray}6.1e+06 & \cellcolor{lightgray}2.8e+08 & \cellcolor{lightgray}2.2e+07 & \cellcolor{lightgray}1.7e+06 & \cellcolor{lightgray}1.0e+05 & \cellcolor{lightgray}9.6e+04 & \cellcolor{lightgray}3.5e+05 \\
\textbf{} & \textbf{max} & 2.7e+07 & 2.4e+07 & 3.5e+07 & 3.1e+08 & 9.0e+07 & 5.1e+06 & 7.8e+06 & 1.8e+06 & 3.1e+06 \\
\cline{1-11}
\multirow[t]{3}{*}{\textbf{Age}} & \textbf{min} & 2.7e+04 & 1.0e+04 & 4.2e+05 & 3.3e+04 & 2.4e+04 & 2.7e+04 & 5.3e+03 & 1.5e+04 & 3.5e+03 \\
\textbf{} & \textbf{best} & \cellcolor{lightgray}4.3e+05 & \cellcolor{lightgray}1.2e+05 & \cellcolor{lightgray}9.0e+06 & \cellcolor{lightgray}1.0e+06 & \cellcolor{lightgray}3.6e+05 & \cellcolor{lightgray}2.9e+06 & \cellcolor{lightgray}2.7e+04 & \cellcolor{lightgray}3.9e+04 & \cellcolor{lightgray}5.2e+05 \\
\textbf{} & \textbf{max} & 1.9e+07 & 1.1e+07 & 1.3e+07 & 1.6e+07 & 2.9e+06 & 3.5e+06 & 6.6e+06 & 2.2e+06 & 1.6e+06 \\
\cline{1-11}
\multirow[t]{3}{*}{\textbf{Taobao}} & \textbf{min} & 7.6e+04 & 7.8e+03 & 3.0e+06 & 1.7e+04 & 1.2e+04 & 9.9e+03 & 5.5e+04 & 5.4e+04 & 3.6e+03 \\
\textbf{} & \textbf{best} & \cellcolor{lightgray}4.0e+06 & \cellcolor{lightgray}1.0e+05 & \cellcolor{lightgray}7.4e+06 & \cellcolor{lightgray}1.5e+06 & \cellcolor{lightgray}1.2e+06 & \cellcolor{lightgray}1.4e+06 & \cellcolor{lightgray}4.1e+06 & \cellcolor{lightgray}1.3e+06 & \cellcolor{lightgray}1.1e+05 \\
\textbf{} & \textbf{max} & 2.2e+07 & 1.6e+07 & 3.1e+07 & 1.6e+07 & 3.1e+06 & 3.5e+06 & 1.2e+07 & 2.5e+06 & 1.6e+06 \\
\cline{1-11}
\multirow[t]{3}{*}{\textbf{BPI17}} & \textbf{min} & 9.3e+04 & 2.7e+04 & 8.4e+05 & 1.4e+06 & 7.1e+04 & 8.6e+04 & 2.7e+04 & 1.2e+04 & 7.3e+03 \\
\textbf{} & \textbf{best} & \cellcolor{lightgray}6.1e+05 & \cellcolor{lightgray}2.1e+05 & \cellcolor{lightgray}5.0e+06 & \cellcolor{lightgray}1.6e+08 & \cellcolor{lightgray}1.7e+06 & \cellcolor{lightgray}4.4e+06 & \cellcolor{lightgray}1.5e+05 & \cellcolor{lightgray}1.3e+05 & \cellcolor{lightgray}1.6e+05 \\
\textbf{} & \textbf{max} & 1.7e+07 & 2.5e+07 & 2.2e+07 & 3.5e+08 & 6.1e+07 & 1.7e+07 & 1.0e+07 & 3.7e+06 & 3.5e+06 \\
\cline{1-11}
\multirow[t]{3}{*}{\textbf{PhysioNet2012}} & \textbf{min} & 3.6e+04 & 1.7e+04 & 8.1e+05 & 1.0e+07 & 7.8e+04 & 2.5e+04 & 1.3e+04 & 1.3e+04 & 5.8e+03 \\
\textbf{} & \textbf{best} & \cellcolor{lightgray}4.0e+05 & \cellcolor{lightgray}2.3e+06 & \cellcolor{lightgray}1.7e+06 & \cellcolor{lightgray}3.1e+08 & \cellcolor{lightgray}3.5e+07 & \cellcolor{lightgray}5.4e+06 & \cellcolor{lightgray}7.6e+05 & \cellcolor{lightgray}6.6e+04 & \cellcolor{lightgray}4.7e+04 \\
\textbf{} & \textbf{max} & 2.7e+07 & 3.3e+07 & 3.5e+07 & 1.8e+09 & 5.4e+08 & 2.2e+07 & 1.1e+07 & 3.1e+06 & 5.3e+06 \\
\cline{1-11}
\multirow[t]{3}{*}{\textbf{MIMIC-III}} & \textbf{min} & 8.2e+04 & 7.9e+03 & 5.6e+05 & 4.5e+05 & 8.7e+03 & 6.3e+03 & 5.2e+03 & 1.5e+04 & 5.5e+03 \\
\textbf{} & \textbf{best} & \cellcolor{lightgray}2.4e+05 & \cellcolor{lightgray}1.1e+06 & \cellcolor{lightgray}1.3e+07 & \cellcolor{lightgray}2.2e+07 & \cellcolor{lightgray}8.0e+05 & \cellcolor{lightgray}2.6e+06 & \cellcolor{lightgray}4.9e+04 & \cellcolor{lightgray}1.5e+06 & \cellcolor{lightgray}1.0e+06 \\
\textbf{} & \textbf{max} & 2.2e+07 & 1.7e+07 & 3.3e+07 & 2.5e+08 & 4.5e+07 & 1.4e+07 & 9.6e+06 & 2.8e+06 & 2.9e+06 \\
\cline{1-11}
\multirow[t]{3}{*}{\textbf{Pendulum}} & \textbf{min} & 2.5e+04 & 6.3e+03 & 2.6e+05 & 2.0e+04 & 3.3e+03 & 3.4e+03 & 8.0e+03 & 1.6e+04 & 7.8e+02 \\
\textbf{} & \textbf{best} & \cellcolor{lightgray}3.3e+05 & \cellcolor{lightgray}1.3e+05 & \cellcolor{lightgray}6.5e+06 & \cellcolor{lightgray}5.9e+06 & \cellcolor{lightgray}2.6e+06 & \cellcolor{lightgray}6.2e+05 & \cellcolor{lightgray}1.0e+06 & \cellcolor{lightgray}1.2e+06 & \cellcolor{lightgray}3.9e+04 \\
\textbf{} & \textbf{max} & 2.2e+07 & 2.2e+07 & 1.8e+07 & 1.6e+07 & 2.9e+06 & 6.4e+06 & 1.1e+07 & 2.7e+06 & 1.8e+06 \\
\cline{1-11}
\multirow[t]{3}{*}{\textbf{ArabicDigits}} & \textbf{min} & 3.8e+04 & 3.9e+03 & 3.7e+06 & 8.9e+04 & 1.6e+04 & 1.5e+05 & 1.8e+04 & 3.5e+04 & 3.3e+03 \\
\textbf{} & \textbf{best} & \cellcolor{lightgray}5.5e+06 & \cellcolor{lightgray}5.3e+05 & \cellcolor{lightgray}1.1e+07 & \cellcolor{lightgray}2.5e+05 & \cellcolor{lightgray}5.7e+07 & \cellcolor{lightgray}4.8e+06 & \cellcolor{lightgray}2.1e+05 & \cellcolor{lightgray}6.5e+05 & \cellcolor{lightgray}3.0e+06 \\
\textbf{} & \textbf{max} & 2.4e+07 & 2.5e+07 & 3.7e+07 & 3.5e+08 & 6.1e+07 & 3.0e+07 & 6.7e+06 & 3.9e+06 & 3.5e+06 \\
\cline{1-11}
\multirow[t]{3}{*}{\textbf{ElectricDevices}} & \textbf{min} & 2.5e+04 & 7.2e+03 & 2.4e+06 & 3.8e+04 & 8.0e+02 & 5.1e+03 & 1.3e+04 & 2.1e+03 & 4.4e+02 \\
\textbf{} & \textbf{best} & \cellcolor{lightgray}1.2e+07 & \cellcolor{lightgray}3.6e+05 & \cellcolor{lightgray}1.7e+07 & \cellcolor{lightgray}3.4e+05 & \cellcolor{lightgray}2.9e+05 & \cellcolor{lightgray}2.0e+06 & \cellcolor{lightgray}4.4e+04 & \cellcolor{lightgray}9.6e+04 & \cellcolor{lightgray}7.3e+05 \\
\textbf{} & \textbf{max} & 2.2e+07 & 2.2e+07 & 4.0e+07 & 7.1e+06 & 1.3e+06 & 4.3e+06 & 8.2e+06 & 2.1e+06 & 1.3e+06 \\
\cline{1-11}
\bottomrule
\end{tabular}
}
\label{tab:learnable_parameters}
\end{table*}

\begin{table*}[t]
\caption{Including vs. Excluding time as a feature. We compare the top 5 test metrics from the \textbf{HPO step} for each option and report the relative performance difference of metrics obtained excluding time feature to metrics obtained including time as a feature. MLEM not included since it has fixed time process option - copied from best CoLES.}
\label{tab:time_to_none}
\centering
\renewcommand{\arraystretch}{1.2}
\resizebox{\textwidth}{!}{%
\begin{tabular}{r|ccccc|ccc|cc}
\toprule
Category & \multicolumn{5}{c|}{Discrete \ES} & \multicolumn{3}{c|}{Continuous \ES} & \multicolumn{2}{c|}{Time Series} \\
\midrule
Dataset & \textbf{MBD} & \textbf{Retail} & \textbf{Age} & \textbf{Taobao} & \textbf{BPI17} & \textbf{PhysioNet2012} & \textbf{MIMIC-III} & \textbf{Pendulum} & \textbf{ArabicDigits} & \textbf{ElectricDevices} \\
\footnotesize{Metric} & \footnotesize{Mean ROC AUC} & \footnotesize{Accuracy} & \footnotesize{Accuracy} & \footnotesize{ROC AUC} & \footnotesize{ROC AUC} & \footnotesize{ROC AUC} & \footnotesize{ROC AUC} & \footnotesize{Accuracy} & \footnotesize{Accuracy} & \footnotesize{Accuracy} \\
\midrule
\textbf{CoLES} & \cellcolor{gray!50}$-0.99 \%$ & $0.18 \%$ & \cellcolor{gray!15}$-0.31 \%$ & \cellcolor{gray!50}$-1.32 \%$ & $-0.16 \%$ & \cellcolor{gray!15}$-0.41 \%$ & \cellcolor{gray!15}$-0.46 \%$ & \cellcolor{gray!100}$-61.34 \%$ & \cellcolor{gray!15}$-0.38 \%$ & \cellcolor{gray!50}$-1.75 \%$ \\
\textbf{GRU} & \cellcolor{gray!50}$-1.17 \%$ & $0.12 \%$ & \cellcolor{gray!25}$-0.82 \%$ & \cellcolor{gray!50}$-1.93 \%$ & \cellcolor{gray!15}$-0.34 \%$ & $-0.27 \%$ & \cellcolor{gray!15}$-0.43 \%$ & \cellcolor{gray!100}$-60.40 \%$ & $-0.08 \%$ & \cellcolor{gray!25}$-0.55 \%$ \\
\textbf{Transformer} & \cellcolor{gray!50}$-1.36 \%$ & \cellcolor{gray!75}$-2.08 \%$ & \cellcolor{gray!25}$-0.84 \%$ & \cellcolor{gray!50}$-1.22 \%$ & $0.40 \%$ & \cellcolor{gray!25}$-0.84 \%$ & \cellcolor{gray!25}$-0.88 \%$ & \cellcolor{gray!100}$-56.99 \%$ & $-0.10 \%$ & $0.82 \%$ \\
\textbf{Mamba} & \cellcolor{gray!50}$-0.99 \%$ & $-0.02 \%$ & \cellcolor{gray!25}$-0.65 \%$ & \cellcolor{gray!50}$-1.65 \%$ & $0.53 \%$ & \cellcolor{gray!25}$-0.57 \%$ & \cellcolor{gray!25}$-0.57 \%$ & \cellcolor{gray!100}$-59.31 \%$ & $-0.13 \%$ & \cellcolor{gray!25}$-0.73 \%$ \\
\textbf{ConvTran} & \cellcolor{gray!25}$-0.62 \%$ & \cellcolor{gray!25}$-0.57 \%$ & \cellcolor{gray!25}$-0.53 \%$ & $-0.09 \%$ & $-0.16 \%$ & \cellcolor{gray!15}$-0.39 \%$ & \cellcolor{gray!15}$-0.49 \%$ & \cellcolor{gray!100}$-59.16 \%$ & $-0.09 \%$ & \cellcolor{gray!75}$-3.84 \%$ \\
\textbf{mTAND} & \cellcolor{gray!75}$-9.67 \%$ & \cellcolor{gray!15}$-0.42 \%$ & \cellcolor{gray!25}$-0.63 \%$ & \cellcolor{gray!75}$-3.14 \%$ & $-0.22 \%$ & \cellcolor{gray!15}$-0.38 \%$ & \cellcolor{gray!25}$-0.81 \%$ & \cellcolor{gray!100}$-22.87 \%$ & \cellcolor{gray!50}$-1.37 \%$ & $0.08 \%$ \\
\textbf{PrimeNet} & \cellcolor{gray!75}$-4.62 \%$ & \cellcolor{gray!50}$-1.03 \%$ & \cellcolor{gray!50}$-1.02 \%$ & \cellcolor{gray!75}$-4.45 \%$ & \cellcolor{gray!25}$-0.51 \%$ & $-0.09 \%$ & \cellcolor{gray!25}$-0.54 \%$ & \cellcolor{gray!100}$-14.26 \%$ & \cellcolor{gray!25}$-0.68 \%$ & \cellcolor{gray!75}$-2.78 \%$ \\
\textbf{MLP} & \cellcolor{gray!50}$-1.03 \%$ & $-0.16 \%$ & \cellcolor{gray!50}$-0.98 \%$ & \cellcolor{gray!75}$-6.44 \%$ & $0.00 \%$ & $-0.12 \%$ & \cellcolor{gray!50}$-1.18 \%$ & \cellcolor{gray!100}$-10.72 \%$ & \cellcolor{gray!75}$-6.91 \%$ & \cellcolor{gray!50}$-1.54 \%$ \\
\bottomrule
\end{tabular}

}
\end{table*}

\begin{table*}
\centering
\caption{Different aggregation approaches: mean across all hidden states or last hidden state. We take top 5 test metrics from \textbf{HPO step} for each option and report mean and std. Highlighted bold if adding time significantly improves performance.}
\label{tab:aggregation}
\resizebox{\textwidth}{!}{\begin{tabular}{rrrrrrrrr}
\toprule
 &  & CoLES & GRU & MLEM & Transformer & Mamba & mTAND & MLP \\
Dataset & Aggregation &  &  &  &  &  &  &  \\
\midrule
\multirow[t]{2}{*}{\textbf{MBD}} & \textbf{Last hidden} & $\cellcolor{lightgray}\bm{0.825 \pm 0.000}$ & $\cellcolor{lightgray}\bm{0.826 \pm 0.000}$ & $\cellcolor{lightgray}\bm{0.823 \pm 0.001}$ & $0.818 \pm 0.001$ & $0.822 \pm 0.001$ & $\cellcolor{lightgray}\bm{0.795 \pm 0.002}$ & $0.755 \pm 0.001$ \\
\textbf{} & \textbf{Mean hidden} & $0.820 \pm 0.002$ & $0.822 \pm 0.001$ & $0.816 \pm 0.006$ & $\cellcolor{lightgray}\bm{0.822 \pm 0.001}$ & $0.822 \pm 0.001$ & $0.786 \pm 0.002$ & $\cellcolor{lightgray}\bm{0.809 \pm 0.000}$ \\
\cline{1-9}
\multirow[t]{2}{*}{\textbf{Retail}} & \textbf{Last hidden} & $\cellcolor{lightgray}\bm{0.551 \pm 0.000}$ & $\cellcolor{lightgray}\bm{0.544 \pm 0.001}$ & $\cellcolor{lightgray}\bm{0.546 \pm 0.001}$ & $0.537 \pm 0.001$ & $0.528 \pm 0.001$ & $0.518 \pm 0.001$ & $0.343 \pm 0.000$ \\
\textbf{} & \textbf{Mean hidden} & $0.547 \pm 0.001$ & $0.542 \pm 0.001$ & $0.539 \pm 0.002$ & $0.541 \pm 0.002$ & $\cellcolor{lightgray}\bm{0.540 \pm 0.000}$ & $0.519 \pm 0.001$ & $\cellcolor{lightgray}\bm{0.526 \pm 0.000}$ \\
\cline{1-9}
\multirow[t]{2}{*}{\textbf{Age}} & \textbf{Last hidden} & $\cellcolor{lightgray}\bm{0.641 \pm 0.001}$ & $0.620 \pm 0.001$ & $\cellcolor{lightgray}\bm{0.647 \pm 0.001}$ & $0.603 \pm 0.004$ & $0.588 \pm 0.007$ & $\cellcolor{lightgray}\bm{0.589 \pm 0.002}$ & $0.338 \pm 0.002$ \\
\textbf{} & \textbf{Mean hidden} & $0.638 \pm 0.001$ & $\cellcolor{lightgray}\bm{0.630 \pm 0.001}$ & $0.635 \pm 0.002$ & $\cellcolor{lightgray}\bm{0.624 \pm 0.002}$ & $\cellcolor{lightgray}\bm{0.620 \pm 0.002}$ & $0.582 \pm 0.004$ & $\cellcolor{lightgray}\bm{0.597 \pm 0.004}$ \\
\cline{1-9}
\multirow[t]{2}{*}{\textbf{Taobao}} & \textbf{Last hidden} & $\cellcolor{lightgray}\bm{0.718 \pm 0.001}$ & $\cellcolor{lightgray}\bm{0.718 \pm 0.001}$ & $\cellcolor{lightgray}\bm{0.718 \pm 0.000}$ & $0.712 \pm 0.001$ & $0.680 \pm 0.012$ & $\cellcolor{lightgray}\bm{0.680 \pm 0.001}$ & $0.603 \pm 0.023$ \\
\textbf{} & \textbf{Mean hidden} & $0.710 \pm 0.001$ & $0.713 \pm 0.001$ & $0.709 \pm 0.001$ & $0.712 \pm 0.001$ & $\cellcolor{lightgray}\bm{0.710 \pm 0.002}$ & $0.676 \pm 0.001$ & $\cellcolor{lightgray}\bm{0.685 \pm 0.001}$ \\
\cline{1-9}
\multirow[t]{2}{*}{\textbf{BPI17}} & \textbf{Last hidden} & $\cellcolor{lightgray}\bm{0.759 \pm 0.002}$ & $\cellcolor{lightgray}\bm{0.760 \pm 0.001}$ & $\cellcolor{lightgray}\bm{0.762 \pm 0.001}$ & $0.758 \pm 0.001$ & $0.741 \pm 0.002$ & $0.733 \pm 0.002$ & $0.709 \pm 0.004$ \\
\textbf{} & \textbf{Mean hidden} & $0.737 \pm 0.006$ & $0.755 \pm 0.002$ & $0.747 \pm 0.002$ & $0.757 \pm 0.002$ & $\cellcolor{lightgray}\bm{0.749 \pm 0.002}$ & $\cellcolor{lightgray}\bm{0.740 \pm 0.002}$ & $\cellcolor{lightgray}\bm{0.738 \pm 0.001}$ \\
\cline{1-9}
\multirow[t]{2}{*}{\textbf{PhysioNet2012}} & \textbf{Last hidden} & $\cellcolor{lightgray}\bm{0.846 \pm 0.001}$ & $\cellcolor{lightgray}\bm{0.848 \pm 0.001}$ & $\cellcolor{lightgray}\bm{0.849 \pm 0.002}$ & $\cellcolor{lightgray}\bm{0.843 \pm 0.001}$ & $\cellcolor{lightgray}\bm{0.840 \pm 0.002}$ & $0.844 \pm 0.000$ & $\cellcolor{lightgray}\bm{0.845 \pm 0.001}$ \\
\textbf{} & \textbf{Mean hidden} & $0.827 \pm 0.002$ & $0.818 \pm 0.003$ & $0.831 \pm 0.003$ & $0.836 \pm 0.005$ & $0.809 \pm 0.007$ & $0.845 \pm 0.001$ & $0.815 \pm 0.003$ \\
\cline{1-9}
\multirow[t]{2}{*}{\textbf{MIMIC-III}} & \textbf{Last hidden} & $\cellcolor{lightgray}\bm{0.908 \pm 0.001}$ & $\cellcolor{lightgray}\bm{0.901 \pm 0.001}$ & $\cellcolor{lightgray}\bm{0.899 \pm 0.000}$ & $\cellcolor{lightgray}\bm{0.894 \pm 0.001}$ & $0.887 \pm 0.003$ & $\cellcolor{lightgray}\bm{0.891 \pm 0.002}$ & $\cellcolor{lightgray}\bm{0.880 \pm 0.000}$ \\
\textbf{} & \textbf{Mean hidden} & $0.898 \pm 0.001$ & $0.894 \pm 0.001$ & $0.896 \pm 0.001$ & $0.890 \pm 0.001$ & $\cellcolor{lightgray}\bm{0.898 \pm 0.001}$ & $0.886 \pm 0.001$ & $0.874 \pm 0.001$ \\
\cline{1-9}
\multirow[t]{2}{*}{\textbf{Pendulum}} & \textbf{Last hidden} & $0.725 \pm 0.007$ & $0.671 \pm 0.011$ & $0.677 \pm 0.008$ & $\cellcolor{lightgray}\bm{0.633 \pm 0.007}$ & $0.640 \pm 0.005$ & $\cellcolor{lightgray}\bm{0.780 \pm 0.012}$ & $\cellcolor{lightgray}\bm{0.194 \pm 0.000}$ \\
\textbf{} & \textbf{Mean hidden} & $0.725 \pm 0.002$ & $\cellcolor{lightgray}\bm{0.703 \pm 0.009}$ & $0.665 \pm 0.005$ & $0.580 \pm 0.022$ & $\cellcolor{lightgray}\bm{0.681 \pm 0.002}$ & $0.711 \pm 0.007$ & $0.158 \pm 0.002$ \\
\cline{1-9}
\multirow[t]{2}{*}{\textbf{ArabicDigits}} & \textbf{Last hidden} & $0.992 \pm 0.001$ & $0.991 \pm 0.002$ & $0.994 \pm 0.001$ & $0.985 \pm 0.001$ & $0.989 \pm 0.001$ & $0.964 \pm 0.002$ & $0.477 \pm 0.007$ \\
\textbf{} & \textbf{Mean hidden} & $0.990 \pm 0.002$ & $0.992 \pm 0.001$ & $0.995 \pm 0.000$ & $\cellcolor{lightgray}\bm{0.991 \pm 0.000}$ & $\cellcolor{lightgray}\bm{0.993 \pm 0.001}$ & $0.958 \pm 0.005$ & $\cellcolor{lightgray}\bm{0.775 \pm 0.001}$ \\
\cline{1-9}
\multirow[t]{2}{*}{\textbf{ElectricDevices}} & \textbf{Last hidden} & $\cellcolor{lightgray}\bm{0.758 \pm 0.005}$ & $0.752 \pm 0.002$ & $\cellcolor{lightgray}\bm{0.749 \pm 0.003}$ & $0.728 \pm 0.007$ & $0.729 \pm 0.004$ & $0.640 \pm 0.005$ & $\cellcolor{lightgray}\bm{0.465 \pm 0.001}$ \\
\textbf{} & \textbf{Mean hidden} & $0.737 \pm 0.004$ & $0.749 \pm 0.003$ & $0.732 \pm 0.004$ & $0.731 \pm 0.009$ & $\cellcolor{lightgray}\bm{0.750 \pm 0.003}$ & $0.640 \pm 0.011$ & $0.254 \pm 0.000$ \\
\cline{1-9}
\bottomrule
\end{tabular}

}
\end{table*}

\begin{table*}
\centering
\caption{Different normalization approaches: with vs without Batch Normalization for input features. We compare the top 5 test metrics from the \textbf{HPO step} for each option and report the relative performance difference of metrics obtained without batch normalization to metrics obtained with batch normalization.}
\label{tab:normalization}
\resizebox{\textwidth}{!}{
\begin{tabular}{r|ccccc|ccc|cc}
\toprule
Category & \multicolumn{5}{c|}{Discrete \ES} & \multicolumn{3}{c|}{Continuous \ES} & \multicolumn{2}{c|}{Time Series} \\
\midrule
Dataset & \textbf{MBD} & \textbf{Retail} & \textbf{Age} & \textbf{Taobao} & \textbf{BPI17} & \textbf{PhysioNet2012} & \textbf{MIMIC-III} & \textbf{Pendulum} & \textbf{ArabicDigits} & \textbf{ElectricDevices} \\
\footnotesize{Metric} & \footnotesize{Mean ROC AUC} & \footnotesize{Accuracy} & \footnotesize{Accuracy} & \footnotesize{ROC AUC} & \footnotesize{ROC AUC} & \footnotesize{ROC AUC} & \footnotesize{ROC AUC} & \footnotesize{Accuracy} & \footnotesize{Accuracy} & \footnotesize{Accuracy} \\
\midrule
\textbf{CoLES} & \cellcolor{gray!15}$-0.25 \%$ & \cellcolor{gray!75}$-3.30 \%$ & \cellcolor{gray!25}$-0.68 \%$ & \cellcolor{gray!50}$-1.36 \%$ & \cellcolor{gray!75}$-3.53 \%$ & \cellcolor{gray!100}$-7.28 \%$ & \cellcolor{gray!75}$-2.54 \%$ & $11.78 \%$ & \cellcolor{gray!15}$-0.06 \%$ & \cellcolor{gray!15}$-0.46 \%$ \\
\textbf{GRU} & \cellcolor{gray!25}$-0.57 \%$ & \cellcolor{gray!75}$-4.01 \%$ & \cellcolor{gray!15}$-0.30 \%$ & \cellcolor{gray!50}$-1.32 \%$ & \cellcolor{gray!75}$-3.83 \%$ & \cellcolor{gray!100}$-7.75 \%$ & \cellcolor{gray!75}$-2.53 \%$ & $10.94 \%$ & $0.06 \%$ & \cellcolor{gray!15}$-0.42 \%$ \\
\textbf{MLEM} & $5.82 \%$ & \cellcolor{gray!75}$-2.95 \%$ & $0.30 \%$ & \cellcolor{gray!50}$-1.21 \%$ & \cellcolor{gray!75}$-4.14 \%$ & \cellcolor{gray!100}$-12.46 \%$ & \cellcolor{gray!75}$-2.39 \%$ & $16.94 \%$ & $0.15 \%$ & \cellcolor{gray!50}$-1.36 \%$ \\
\textbf{Transformer} & $0.00 \%$ & \cellcolor{gray!100}$-21.29 \%$ & \cellcolor{gray!15}$-0.10 \%$ & \cellcolor{gray!25}$-0.68 \%$ & \cellcolor{gray!100}$-5.59 \%$ & \cellcolor{gray!100}$-6.65 \%$ & \cellcolor{gray!75}$-2.38 \%$ & $15.07 \%$ & \cellcolor{gray!15}$-0.25 \%$ & \cellcolor{gray!50}$-1.47 \%$ \\
\textbf{Mamba} & \cellcolor{gray!15}$-0.46 \%$ & \cellcolor{gray!75}$-3.66 \%$ & \cellcolor{gray!25}$-0.61 \%$ & \cellcolor{gray!50}$-1.27 \%$ & \cellcolor{gray!50}$-1.87 \%$ & \cellcolor{gray!25}$-0.73 \%$ & \cellcolor{gray!50}$-1.38 \%$ & $10.19 \%$ & $0.46 \%$ & $1.05 \%$ \\
\textbf{ConvTran} & $0.07 \%$ & $0.19 \%$ & \cellcolor{gray!75}$-2.31 \%$ & $0.07 \%$ & $0.23 \%$ & \cellcolor{gray!75}$-2.48 \%$ & \cellcolor{gray!25}$-0.61 \%$ & $9.47 \%$ & $0.19 \%$ & \cellcolor{gray!75}$-2.66 \%$ \\
\textbf{mTAND} & $1.66 \%$ & \cellcolor{gray!25}$-0.98 \%$ & \cellcolor{gray!25}$-0.75 \%$ & \cellcolor{gray!75}$-3.61 \%$ & \cellcolor{gray!25}$-0.93 \%$ & \cellcolor{gray!75}$-4.47 \%$ & \cellcolor{gray!50}$-1.46 \%$ & $17.35 \%$ & $0.89 \%$ & $4.24 \%$ \\
\textbf{PrimeNet} & \cellcolor{gray!15}$-0.04 \%$ & \cellcolor{gray!50}$-1.04 \%$ & $0.30 \%$ & \cellcolor{gray!75}$-4.27 \%$ & \cellcolor{gray!50}$-1.08 \%$ & \cellcolor{gray!50}$-1.73 \%$ & $0.09 \%$ & $19.08 \%$ & \cellcolor{gray!25}$-0.91 \%$ & \cellcolor{gray!75}$-2.87 \%$ \\
\textbf{MLP} & \cellcolor{gray!15}$-0.40 \%$ & \cellcolor{gray!75}$-3.00 \%$ & \cellcolor{gray!50}$-1.07 \%$ & \cellcolor{gray!100}$-7.75 \%$ & \cellcolor{gray!50}$-1.03 \%$ & \cellcolor{gray!75}$-4.07 \%$ & \cellcolor{gray!75}$-3.87 \%$ & $5.75 \%$ & \cellcolor{gray!50}$-1.84 \%$ & \cellcolor{gray!15}$-0.18 \%$ \\
\bottomrule
\end{tabular}
}
\end{table*}

\begin{table*}
\caption{Learning Rate Importance by Optuna Ranking (Smaller Rank = Higher Importance). There is a unique best Learning Rate for each Dataset/Method combination}
\label{tab:LR}
\centering
\resizebox{\textwidth}{!}{
\begin{tabular}{r|ccccc|ccc|cc}
\toprule
Category & \multicolumn{5}{c|}{Discrete \ES} & \multicolumn{3}{c|}{Continuous \ES} & \multicolumn{2}{c|}{Time Series} \\
\midrule
Dataset & \textbf{MBD} & \textbf{Retail} & \textbf{Age} & \textbf{Taobao} & \textbf{BPI17} & \textbf{PhysioNet2012} & \textbf{MIMIC-III} & \textbf{Pendulum} & \textbf{ArabicDigits} & \textbf{ElectricDevices} \\
\footnotesize{Metric} & \footnotesize{Mean ROC AUC} & \footnotesize{Accuracy} & \footnotesize{Accuracy} & \footnotesize{ROC AUC} & \footnotesize{ROC AUC} & \footnotesize{ROC AUC} & \footnotesize{ROC AUC} & \footnotesize{Accuracy} & \footnotesize{Accuracy} & \footnotesize{Accuracy} \\
\midrule
\textbf{CoLES} & $9$ & \cellcolor{gray!25}$4$ & \cellcolor{gray!75}$1$ & \cellcolor{gray!75}$1$ & \cellcolor{gray!50}$2$ & \cellcolor{gray!25}$3$ & \cellcolor{gray!50}$2$ & \cellcolor{gray!50}$2$ & \cellcolor{gray!50}$2$ & \cellcolor{gray!75}$1$ \\
\textbf{GRU} & \cellcolor{gray!75}$1$ & \cellcolor{gray!75}$1$ & \cellcolor{gray!50}$2$ & \cellcolor{gray!75}$1$ & \cellcolor{gray!75}$1$ & \cellcolor{gray!25}$4$ & \cellcolor{gray!50}$2$ & \cellcolor{gray!25}$4$ & \cellcolor{gray!75}$1$ & \cellcolor{gray!75}$1$ \\
\textbf{MLEM} & \cellcolor{gray!25}$3$ & \cellcolor{gray!75}$1$ & \cellcolor{gray!50}$2$ & \cellcolor{gray!75}$1$ & \cellcolor{gray!75}$1$ & \cellcolor{gray!25}$4$ & \cellcolor{gray!25}$4$ & $9$ & \cellcolor{gray!75}$1$ & \cellcolor{gray!75}$1$ \\
\textbf{Transformer} & \cellcolor{gray!25}$4$ & $8$ & \cellcolor{gray!75}$1$ & \cellcolor{gray!25}$3$ & \cellcolor{gray!25}$3$ & $7$ & $6$ & $7$ & $5$ & \cellcolor{gray!75}$1$ \\
\textbf{Mamba} & \cellcolor{gray!75}$1$ & \cellcolor{gray!75}$1$ & \cellcolor{gray!75}$1$ & \cellcolor{gray!75}$1$ & \cellcolor{gray!75}$1$ & \cellcolor{gray!75}$1$ & \cellcolor{gray!75}$1$ & \cellcolor{gray!75}$1$ & \cellcolor{gray!75}$1$ & \cellcolor{gray!75}$1$ \\
\textbf{ConvTran} & $5$ & \cellcolor{gray!75}$1$ & $9$ & \cellcolor{gray!50}$2$ & \cellcolor{gray!75}$1$ & \cellcolor{gray!75}$1$ & \cellcolor{gray!75}$1$ & \cellcolor{gray!25}$3$ & $5$ & \cellcolor{gray!50}$2$ \\
\textbf{mTAND} & \cellcolor{gray!75}$1$ & \cellcolor{gray!75}$1$ & \cellcolor{gray!75}$1$ & \cellcolor{gray!75}$1$ & \cellcolor{gray!75}$1$ & \cellcolor{gray!25}$3$ & \cellcolor{gray!75}$1$ & \cellcolor{gray!75}$1$ & \cellcolor{gray!75}$1$ & \cellcolor{gray!75}$1$ \\
\textbf{PrimeNet} & \cellcolor{gray!75}$1$ & \cellcolor{gray!75}$1$ & $5$ & \cellcolor{gray!75}$1$ & \cellcolor{gray!75}$1$ & \cellcolor{gray!50}$2$ & \cellcolor{gray!75}$1$ & \cellcolor{gray!75}$1$ & \cellcolor{gray!75}$1$ & $7$ \\
\textbf{MLP} & \cellcolor{gray!75}$1$ & \cellcolor{gray!75}$1$ & \cellcolor{gray!50}$2$ & \cellcolor{gray!50}$2$ & \cellcolor{gray!75}$1$ & \cellcolor{gray!75}$1$ & \cellcolor{gray!25}$3$ & \cellcolor{gray!75}$1$ & \cellcolor{gray!50}$2$ & \cellcolor{gray!50}$2$ \\
\bottomrule
\end{tabular}
}
\end{table*}




\newpage

\section{Subsets metric relationships} \label{appendix: hpo_corr}

\begin{figure*}[p]
    \centering
    \includegraphics[width=0.7\textwidth]{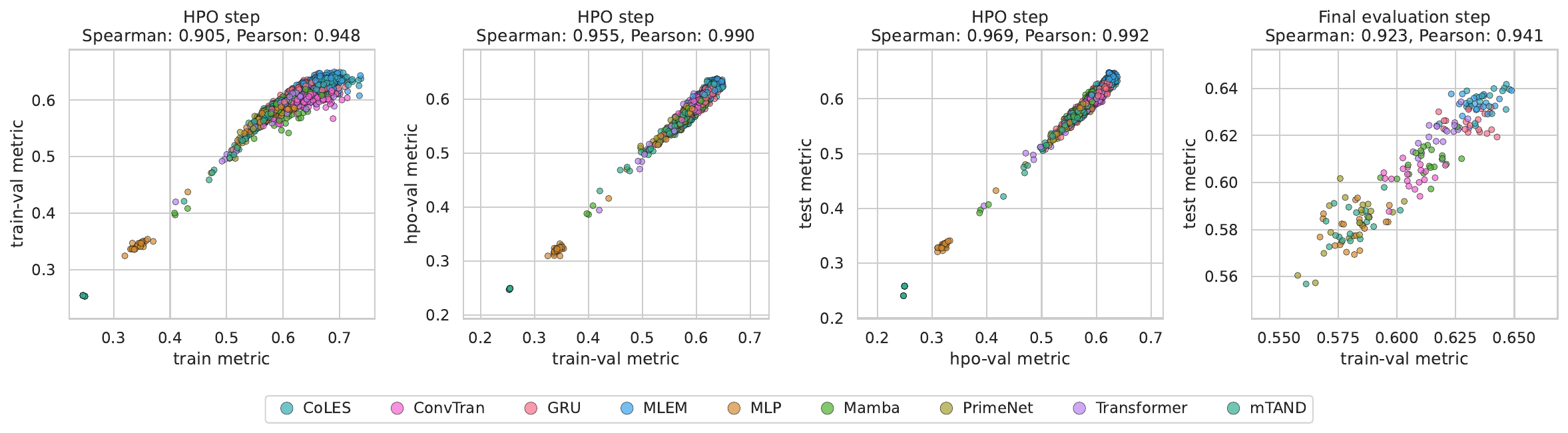}
    \caption{Performance metric relationships and correlations of different subsets among all methods on Age dataset}
    \label{fig:age_hpo_corr}
\end{figure*}

\begin{figure*}[p]
    \centering
    \includegraphics[width=0.7\textwidth]{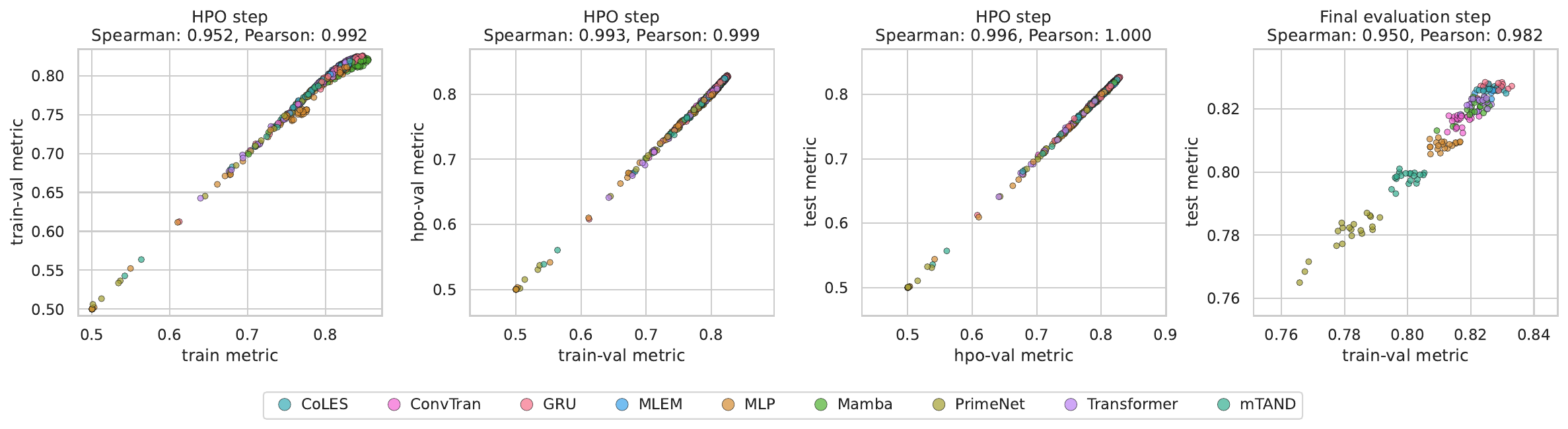}
    \caption{Performance metric relationships and correlations of different subsets among all methods on MBD dataset}
    \label{fig:mbd_hpo_corr}
\end{figure*}

\begin{figure*}[p]
    \centering
    \includegraphics[width=0.7\textwidth]{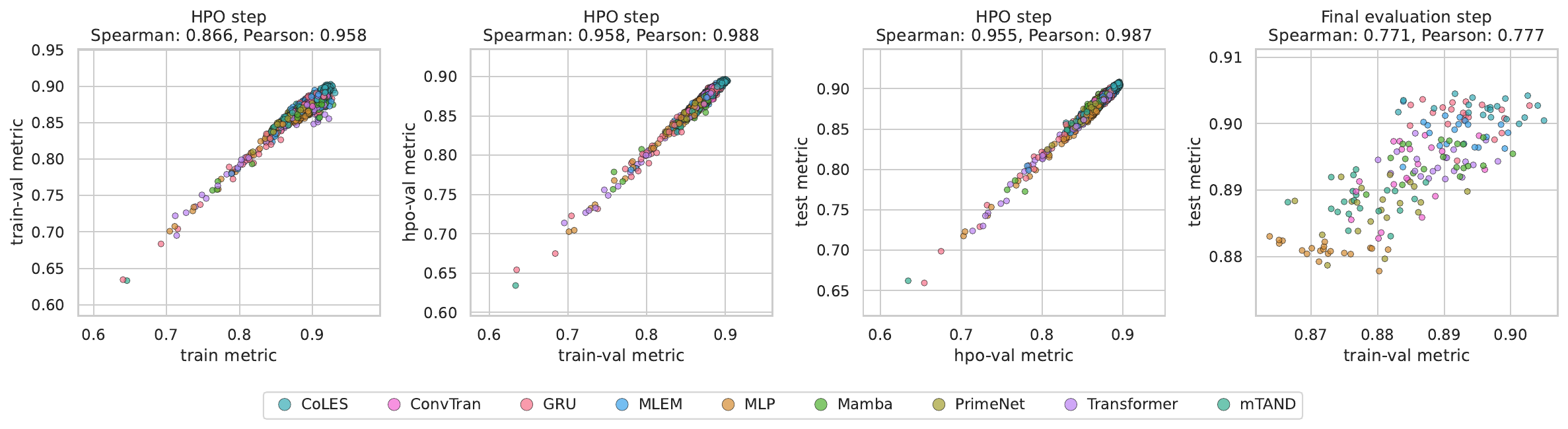}
    \caption{Performance metric relationships and correlations of different subsets among all methods on MIMIC-III dataset}
    \label{fig:mimic3_hpo_corr}
\end{figure*}

\begin{figure*}[p]
    \centering
    \includegraphics[width=0.7\textwidth]{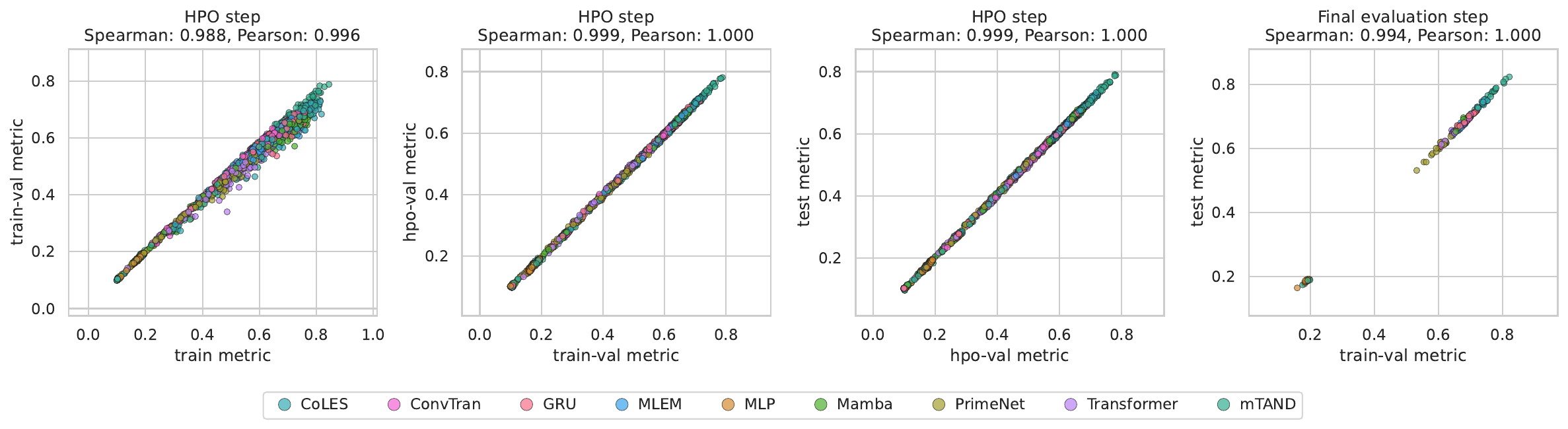}
    \caption{Performance metric relationships and correlations of different subsets among all methods on Pendulum dataset}
    \label{fig:pendulum_hpo_corr}
\end{figure*}

\begin{figure*}[p]
    \centering
    \includegraphics[width=0.7\textwidth]{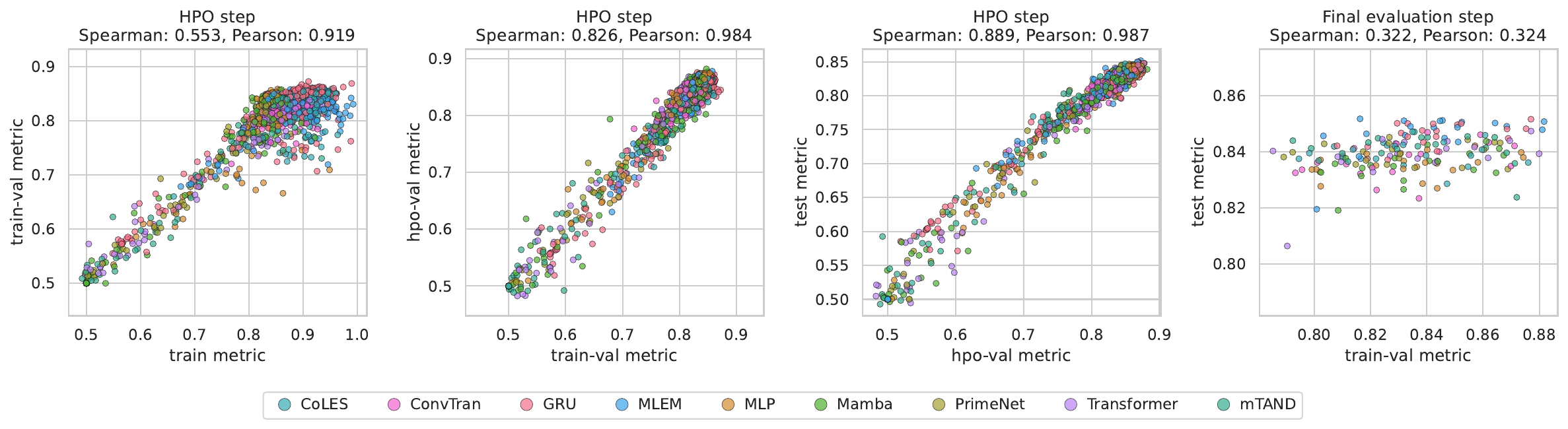}
    \caption{Performance metric relationships and correlations of different subsets among all methods on PhysioNet2012 dataset}
    \label{fig:physionet2012_hpo_corr}
\end{figure*}

\begin{figure*}[p]
    \centering
    \includegraphics[width=0.7\textwidth]{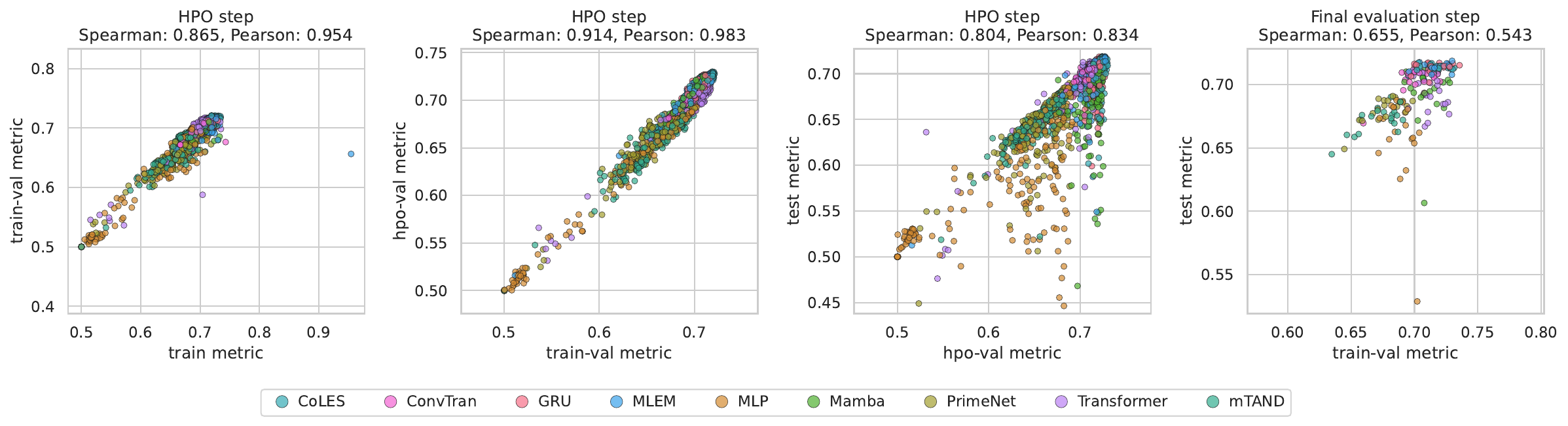}
    \caption{Performance metric relationships and correlations of different subsets among all methods on Taobao dataset}
    \label{fig:taobao_hpo_corr}
\end{figure*}

\begin{figure*}[p]
    \centering
    \includegraphics[width=0.7\textwidth]{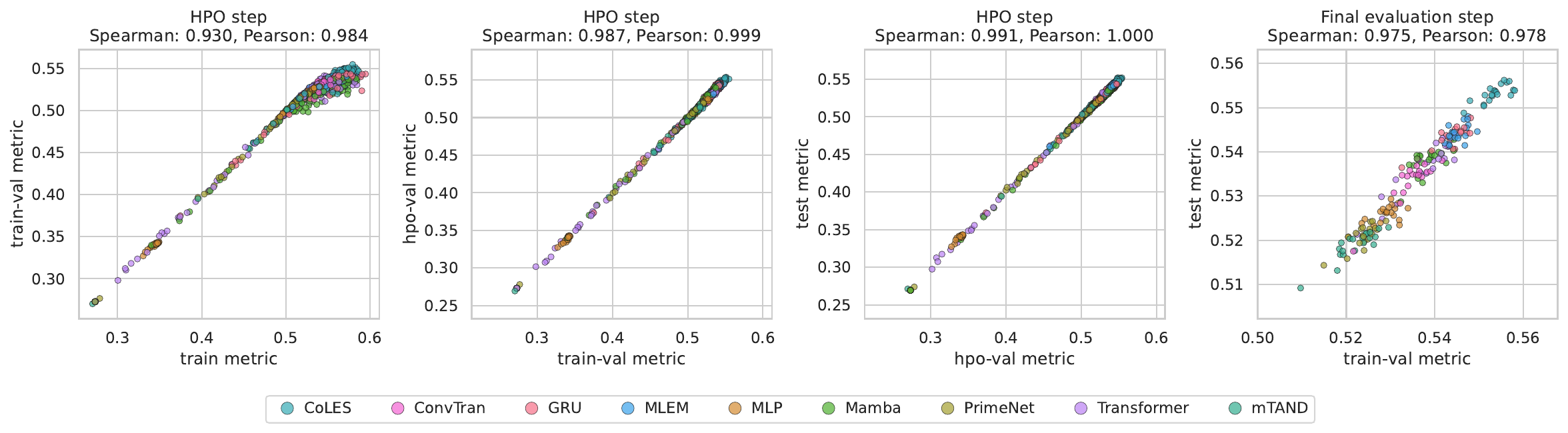}
    \caption{Performance metric relationships and correlations of different subsets among all methods on Retail dataset}
    \label{fig:x5_hpo_corr}
\end{figure*}

\begin{figure*}[p]
    \centering
    \includegraphics[width=0.7\textwidth]{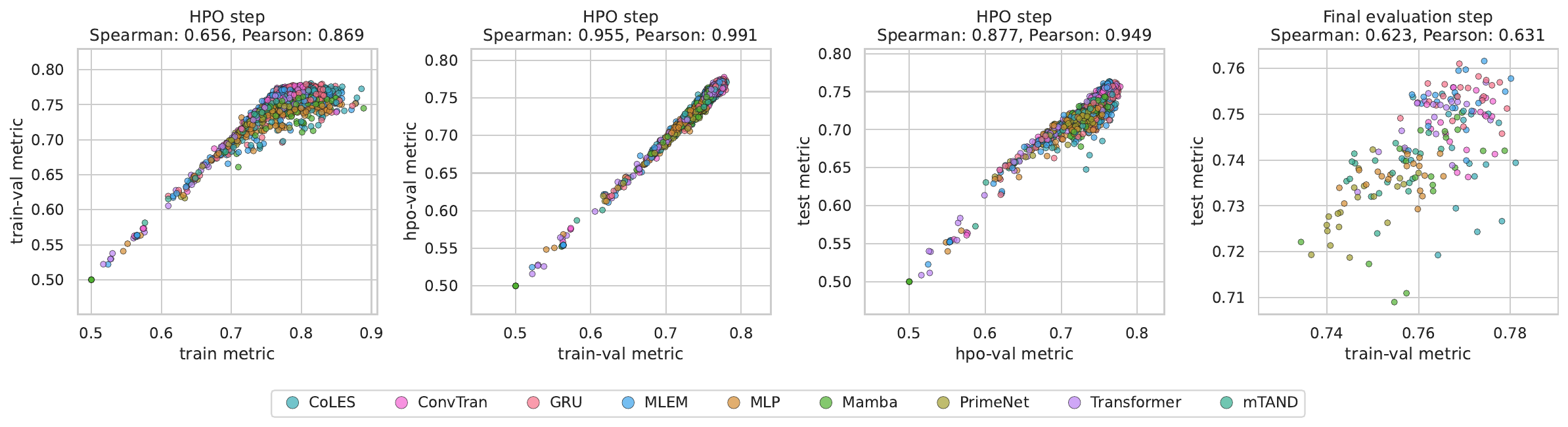}
    \caption{Performance metric relationships and correlations of different subsets among all methods on BPI17 dataset}
    \label{fig:bpi17_hpo_corr}
\end{figure*}

\begin{figure*}[p]
    \centering
    \includegraphics[width=0.7\textwidth]{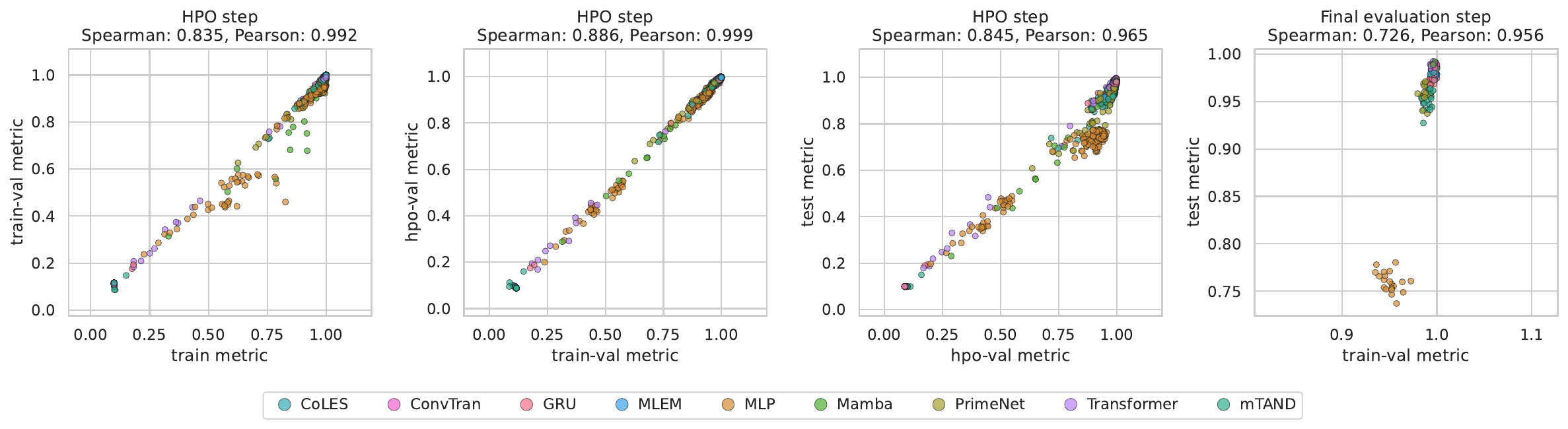}
    \caption{Performance metric relationships and correlations of different subsets among all methods on ArabicDigits dataset}
    \label{fig:arabic_hpo_corr}
\end{figure*}

\begin{figure*}[p]
    \centering
    \includegraphics[width=0.7\textwidth]{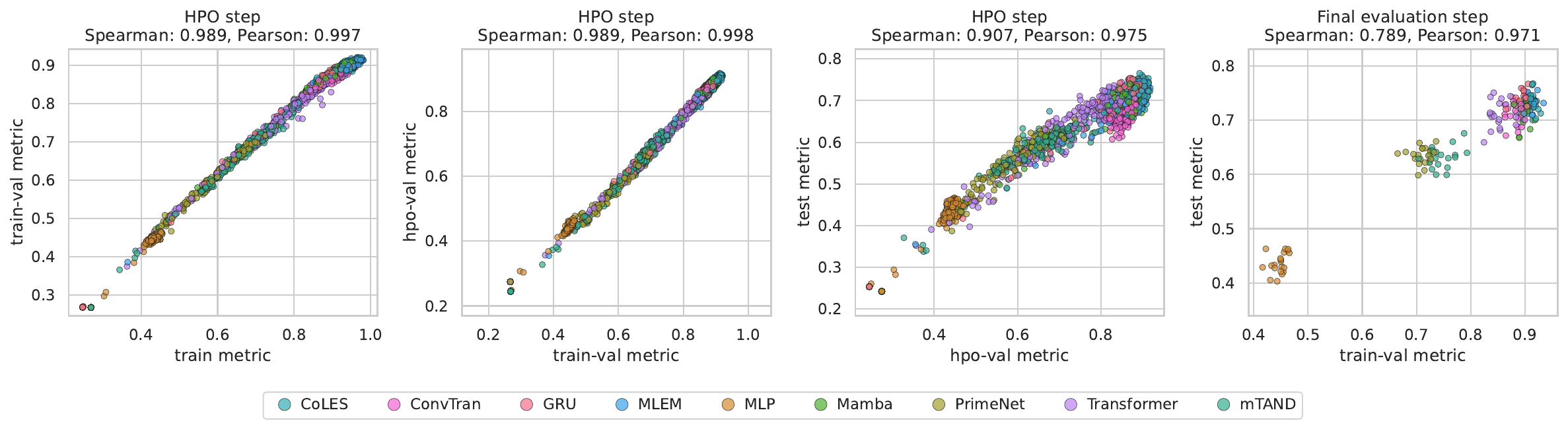}
    \caption{Performance metric relationships and correlations of different subsets among all methods on ElectricDevices dataset}
    \label{fig:electric_devices_hpo_corr}
\end{figure*}
\end{document}